\documentclass{article}

\PassOptionsToPackage{numbers,sort&compress}{natbib}
\usepackage[main, final]{neurips_2026}
\usepackage{graphicx}
\usepackage{wrapfig}

\usepackage{titlesec}
\titlespacing*{\paragraph}{0pt}{0.3ex}{0.3em}

\usepackage[utf8]{inputenc} % allow utf-8 input
\usepackage[T1]{fontenc}    % use 8-bit T1 fonts
\usepackage{hyperref}       % hyperlinks
\usepackage{url}            % simple URL typesetting
\usepackage{booktabs}       % professional-quality tables
\usepackage{amsfonts}       % blackboard math symbols
\usepackage{nicefrac}       % compact symbols for 1/2, etc.
\usepackage{microtype}      % microtypography
\usepackage{xcolor}         % colors
\usepackage{amsmath}
\usepackage{multirow}
\usepackage{mathtools}
\usepackage{listings}
\definecolor{codebg}{RGB}{248,248,248}
\definecolor{codekw}{RGB}{0,92,197}
\definecolor{codecomment}{RGB}{106,115,125}
\definecolor{codestring}{RGB}{3,47,98}
\definecolor{codefunc}{RGB}{111,66,193}
\definecolor{codenum}{RGB}{0,92,197}
\lstdefinestyle{pystyle}{
  language=Python,
  basicstyle=\ttfamily\scriptsize,
  backgroundcolor=\color{codebg},
  keywordstyle=\color{codekw}\bfseries,
  commentstyle=\color{codecomment}\itshape,
  stringstyle=\color{codestring},
  emph={torch,graphdpo_loss,graphdpo_loss_linear},
  emphstyle=\color{codefunc},
  numbers=left, numberstyle=\tiny\color{codecomment}, numbersep=6pt,
  frame=single, rulecolor=\color{black!15}, framesep=4pt, xleftmargin=12pt,
  columns=fullflexible, keepspaces=true, showstringspaces=false,
  breaklines=true, captionpos=t
}

\title{Beyond Pairs: Your Language Model is Secretly Optimizing a Preference Graph}

\author{
Ning Liu \quad Chuanneng Sun \quad Kristina Klinkner \quad Shervin Malmasi \\
Amazon \\
\texttt{\{ningliun,sneng,klinkner,malmasi\}@amazon.com}
}

\begin{document}

\maketitle

\begin{abstract}

Direct Preference Optimization (DPO) aligns language models using pairwise preference comparisons, offering a simple and effective alternative to Reinforcement Learning (RL) from human feedback. However, in many practical settings, training data consists of multiple rollouts per prompt, inducing rich preference structure that pairwise DPO fails to exploit. Collapsing such data into independent pairs discards transitivity, introduces redundant or conflicting supervision, and can lead to unstable optimization. We propose Graph Direct Preference Optimization (GraphDPO), a structured generalization of DPO that operates over directed acyclic preference graphs induced by rollout rankings. GraphDPO encodes dominance relations as edges and optimizes a graph-structured Plackett--Luce-inspired objective that aggregates supervision over dominated neighborhoods, propagating the transitive structure of the input preferences while recovering standard DPO as a special case. GraphDPO naturally handles tied and incomparable responses through an equivalence-class construction: responses with identical preference signals form graph layers, incomparable responses share a layer without an edge, and absent edges contribute zero loss, preventing the spurious gradients that strict total-order objectives incur. Under this layered construction, GraphDPO maintains linear per-prompt complexity via efficient log-sum-exp aggregation. We further incorporate optional ground-truth anchoring by inserting verified solutions as dominant nodes and applying an annealed schedule that stabilizes early training while gradually relaxing oracle supervision. Experiments on reasoning and program synthesis tasks demonstrate superior performance, suggesting that graph-structured preference modeling is a scalable and robust alternative to pairwise and listwise alignment objectives.

\end{abstract}

\section{Introduction}

Aligning large language models (LLMs) with human preferences is essential for deploying safe, reliable, and useful AI systems~\citep{bai2022constitutional, ganguli2022predictability, touvron2023llama}. The dominant paradigm, Reinforcement Learning from Human Feedback (RLHF), trains a reward model from preference comparisons and optimizes the policy using reinforcement learning algorithms such as Proximal Policy Optimization (PPO)~\citep{christiano2017deep, stiennon2020learning, ouyang2022training, schulman2017ppo}. While effective, RLHF pipelines incur substantial costs: they require careful reward model training, sensitive hyperparameter tuning, and repeated on-policy sampling, and remain prone to reward misspecification and optimization instability~\citep{ziegler2019fine,rafailov2023direct, li2023alpacaeval}.

Direct Preference Optimization (DPO)~\citep{rafailov2023direct} offers a major simplification of this pipeline. By reparameterizing the RLHF objective, DPO eliminates explicit reward modeling and directly optimizes policy log-probability ratios against pairwise preference data relative to a fixed reference model. This reformulation reduces alignment to a stable classification-style objective while retaining strong empirical performance. As a result, DPO has inspired a growing family of reference-based, reward-free alignment methods, including IPO~\citep{azar2024general}, KTO~\citep{ethayarajh2024kto}, SimPO~\citep{meng2024simpo}, ORPO~\citep{hong2024orpo}, and iterative or online variants such as SPIN~\citep{chen2024spin}.

\paragraph{The pairwise and listwise bottleneck.}
Despite these advances, most DPO-style methods fundamentally operate on \emph{isolated pairwise comparisons}. This assumption is often mismatched with practical data collection regimes. In many real-world and benchmark settings, preference data is obtained by sampling multiple candidate responses per prompt and ranking them via human judgments, execution-based verification~\citep{hendrycks2021apps}, or structured heuristics~\citep{cobbe2021training, hendrycks2021measuring}. Such multi-sample rankings arise naturally in reasoning and program synthesis tasks, where multiple solution attempts are generated and jointly evaluated~\citep{shao2024deepseekmath,lightman2023lets, wang2022self, liu2026jury}. Standard DPO handles this by decomposing rankings over $K$ responses into independent pairwise comparisons, yielding $O(K^2)$ training instances treated as separate constraints. This reduction introduces several limitations: it discards transitivity, introduces redundant supervision, and fails to exploit global structure present in ranked rollouts~\citep{vinyals2016order, cao2007learning}.

Recent work has begun to explore learning directly from multi-sample rankings. Listwise preference optimization approaches such as PRO~\citep{song2024preference} and LiPO~\citep{liu2025lipo} formulate alignment as learning-to-rank (LTR) over ordered response lists using Plackett--Luce (PL) or LambdaLoss-style objectives~\citep{plackett1975analysis, luce1959individual, burges2010ranknet}. These methods demonstrate that exploiting ranking structure improves alignment performance~\citep{liu2009learning}. However, they fundamentally assume \emph{total orderings}: responses must be arranged into strict ranked lists, and supervision is expressed through listwise permutation likelihoods or weighted pairwise surrogates. In many LLM alignment settings, this assumption is restrictive. Rollout-based evaluation often produces partial orders, equivalence classes under discrete verification signals, or sparse comparability relations that cannot be faithfully represented as strict lists~\citep{wang2024rescue,chen2024preference}.

\paragraph{Graph Direct Preference Optimization.}
We introduce \emph{Graph Direct Preference Optimization} (GraphDPO), a structured generalization of DPO that shifts preference modeling from ordered lists to \emph{preference graphs}. For each prompt, GraphDPO constructs a directed acyclic graph (DAG) over sampled responses that encodes the dominance relations implied by rankings. This formulation unifies existing representations as special cases of one objective: pairwise DPO corresponds to a single-edge graph, chains recover total-order rankings, and tied or incomparable pairs are non-edges within a shared equivalence class, contributing zero gradient and avoiding the spurious supervision that strict listwise models incur~\citep{koller2009probabilistic}. GraphDPO optimizes a graph-structured PL-inspired objective over dominated response neighborhoods: each response is contrasted only against responses in strictly worse preference groups, yielding a groupwise factorization that propagates the transitive structure of the input preferences without requiring total-order supervision. Under the equivalence-class (layered DAG) construction, which all our experiments use, GraphDPO maintains linear complexity in the number of responses: each response contributes a single $\log\sum\exp$ aggregation over dominated samples, avoiding explicit enumeration of $O(K^2)$ pairwise comparisons or normalization over permutations. This yields a scalable objective suitable for rollout-heavy alignment settings.

GraphDPO further supports \emph{ground-truth anchoring} when verified solutions are available: oracle responses are inserted as dominant graph nodes, with an annealed schedule that emphasizes oracle supervision early in training to stabilize optimization and gradually relaxes its influence as the policy improves. We evaluate GraphDPO on GSM8K~\citep{cobbe2021training}, MATH~\citep{hendrycks2021measuring}, and APPS~\citep{hendrycks2021apps}, where it consistently outperforms pairwise DPO, listwise baselines, and rollout-based alignment methods, demonstrating that modeling preference structure as graphs provides a scalable and robust alternative to pairwise and listwise optimization. Our core contributions are:
\begin{itemize}
% \vspace{-0.1in}
\item \textbf{Graph-structured preference optimization.}
We reformulate preference learning from a graph perspective and propose \emph{Graph Direct Preference Optimization} (GraphDPO), a structured generalization of DPO that constructs a preference DAG over multiple rollouts per prompt. GraphDPO leverages the full structure of ranked rollouts rather than treating comparisons as isolated pairs, with pairwise DPO, chains, ties, and incomparability as special cases.
% \vspace{-0.1in}
\item \textbf{Groupwise ranking with tied- and incomparable-preference handling.}
GraphDPO naturally supports equivalence classes of responses with identical preference signals, and handles incomparable responses through the same mechanism: both share a class with no edge between them and contribute zero gradient. Comparisons are performed only across preference groups, avoiding spurious gradients from near-identical samples and enabling robust optimization under discrete or sparse supervision.
% \vspace{-0.0in}
\item \textbf{Linear-time graph learning under the layered construction.}
Under the equivalence-class (layered DAG) construction, GraphDPO maintains linear complexity in the number of responses. Each response contributes a single $\log\sum\exp$ aggregation over dominated samples, capturing global ranking information without enumerating $O(K^2)$ pairwise comparisons; for general partial orders the cost scales as $O(|E|)$. We provide an explicit algorithm and complexity analysis in Appendix~\ref{appd:complexity}.
% \vspace{-0.0in}
\item \textbf{Stabilizing ground-truth anchoring.}
When verified solutions are available, we integrate \emph{offline ground-truth anchoring} into the preference graph by treating ground-truth responses as dominant nodes. An annealed anchoring schedule emphasizes oracle supervision in early training to stabilize optimization, and gradually relaxes its influence to allow greater policy flexibility, improving both convergence and final performance.

\end{itemize}

\section{Related Work}

% \paragraph{Reinforcement Learning from Human Feedback.}
% RLHF trains a reward model from pairwise comparisons and optimizes policies via reinforcement learning algorithms such as PPO~\citep{christiano2017deep, ouyang2022training, schulman2017ppo}. While effective, RLHF introduces substantial complexity, including reward model misspecification, on-policy sampling overhead, and hyperparameter sensitivity~\citep{ziegler2019fine}. Subsequent work has explored offline formulations and reward-free alternatives, motivating the development of direct preference optimization approaches.

\paragraph{Direct Preference Optimization and variants.}
DPO~\citep{rafailov2023direct} removes explicit reward modeling by optimizing log-probability ratios relative to a reference policy. Extensions such as IPO~\citep{azar2024general}, KTO~\citep{ethayarajh2024kto}, SimPO~\citep{meng2024simpo}, ORPO~\citep{hong2024orpo}, and RRHF~\citep{yuan2023rrhf} modify the surrogate objective or feedback assumptions, but continue to treat each comparison independently. Related approaches also include offline preference optimization and margin-based variants~\citep{najafi2026offline, kim2024margin, amini2024direct,pang2026small}. These approaches do not exploit global structure when multiple responses are available per prompt.

\paragraph{Multi-sample preference learning.}
Sampling multiple responses per prompt has been widely used to improve alignment. Best-of-$K$ selection and rejection sampling~\citep{sun2024fast, nakano2021webgpt} filter outputs without learning from relative structure. SPIN~\citep{chen2024spin} iteratively generates and ranks rollouts but ultimately reduces supervision to pairwise updates, while GRPO~\citep{shao2024deepseekmath} relies on reward-based group normalization. Recent methods extend DPO to multi-sample settings: MPO~\citep{gupta2025mpo} performs set-level contrasts over preferred and dispreferred groups, SWEPO~\citep{gupta2025swepo} introduces weighted groupwise objectives, and RPO~\citep{yin2024relative} leverages relative comparisons across prompts. However, these approaches either reduce supervision to pairwise signals, depend on reward estimation or heuristic weighting, or rely on coarse set-level partitions. GraphDPO instead models the structured relations induced by ranked rollouts, capturing fine-grained dependencies without reward modeling or pairwise reduction.

\paragraph{Listwise preference optimization.}
Recent work formulates alignment as a listwise ranking problem. PRO~\citep{song2024preference} applies Plackett--Luce list-MLE objectives to ordered response lists, while LiPO~\citep{liu2025lipo} studies a broader LTR framework and introduces LambdaLoss-weighted objectives~\citep{wang2018lambdaloss} that leverage listwise signals. These methods demonstrate the benefit of optimizing over rankings rather than independent pairs, consistent with findings in the LTR literature~\citep{liu2009learning}. However, they assume strict total orderings and represent supervision as permutations or listwise normalization terms. GraphDPO differs fundamentally by modeling preferences as directed acyclic graphs, enabling partial orders, equivalence classes, and sparse comparability relations that arise in rollout-based evaluation. Furthermore, GraphDPO preserves DPO's reference-ratio formulation and introduces a linear-time dominated-set likelihood factorization rather than permutation-based normalization.

\paragraph{Graph and structured preference modeling.}
Graph-based modeling has been explored in ranking and choice theory~\citep{guiver2009bayesian, dehghani2017neural}, as well as in neural and structured prediction settings~\citep{kipf2017gcn, velivckovic2018gat}. Preference learning with structured or contextual comparisons has also been studied in bandit and ranking settings~\citep{sekhari2023contextual}. Recent work such as PGED~\citep{hu2026towards} highlights the value of preference graphs for enforcing consistency and mitigating contradictory or noisy supervision through graph-based aggregation. However, these ideas have not been integrated into reference-based preference optimization for LLM alignment. GraphDPO bridges this gap by combining graph-structured representations with DPO-style objectives, enabling scalable optimization over structured preference relations while recovering pairwise and listwise formulations as special cases.

\begin{figure*}[!t]\centering
\includegraphics[width=0.99\linewidth]{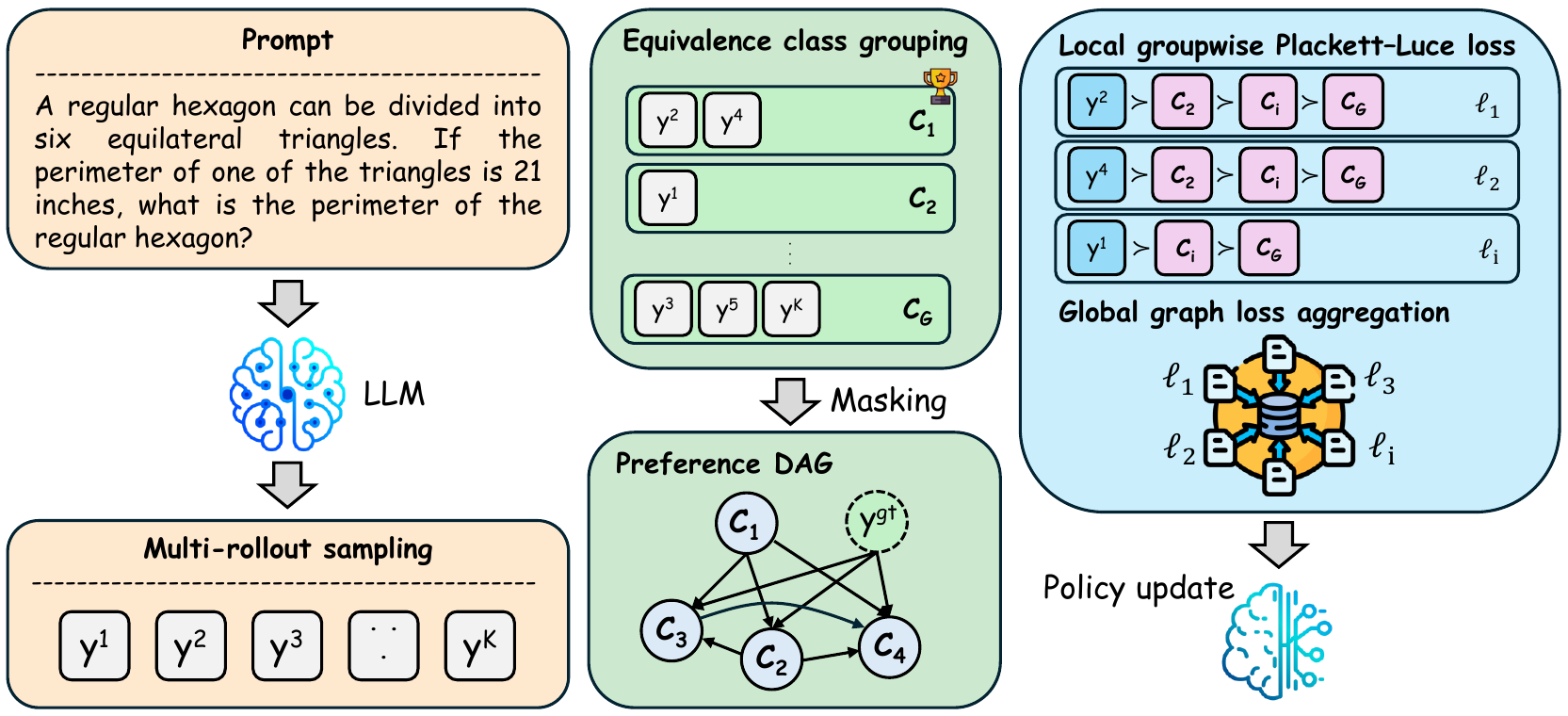}
 \caption{GraphDPO pipeline for LLM alignment. For each prompt, the policy samples $K$ rollouts, which are grouped into equivalence classes according to preference signals. These classes induce a DAG structure whose edges encode dominance relations between groups, with an optional ground-truth node as a global anchor. Equivalence-class masking removes intra-group comparisons so that each response is contrasted only with strictly worse groups via a local Plackett--Luce loss. The resulting losses are aggregated over the graph to update the policy while propagating the transitive structure of the input preferences.}
 \label{fig:architecture}
\end{figure*}

\section{Preliminaries}\label{sec:preliminary}

We consider the problem of aligning a pretrained language model with human or task-specific preferences using relative comparisons. Let $\mathcal{X}$ denote the space of prompts and $\mathcal{Y}$ the space of candidate responses. For a prompt $x \in \mathcal{X}$, a policy $\pi_\theta(y \mid x)$ induces a distribution over responses, where $\theta$ denotes model parameters. Following Direct Preference Optimization (DPO)~\citep{rafailov2023direct}, we assume access to a fixed reference policy $\pi_{\mathrm{ref}}$, typically the pretrained model prior to alignment.

\paragraph{Preference data.}
For each prompt $x$, we observe a finite set of candidate responses $\{y_1, \ldots, y_K\}$ together with an ordinal preference relation $\succ_x$. The relation may be incomplete and allows ties, and therefore defines a partial order rather than a strict ranking. We write $y_i \sim_x y_j$ when neither response is preferred to the other.

This partial order can be represented as equivalence classes
\[
\mathcal{C}_1 \succ_x \cdots \succ_x \mathcal{C}_G,
\]
where responses within a class are tied and only inter-class comparisons are defined. This representation naturally arises when supervision signals are discrete (e.g., correctness indicators).

\paragraph{Direct Preference Optimization.}
DPO derives a tractable objective by reparameterizing KL-regularized reinforcement learning and eliminating the need for an explicit reward model. Under a Bradley--Terry preference model, the probability that $y_i$ is preferred over $y_j$ is
\vspace{-0.0in}\noindent
\begin{equation}
P(y_i \succ y_j \mid x)
= \sigma\bigl(s_\theta(x,y_i)-s_\theta(x,y_j)\bigr),
\end{equation}
\vspace{-0.0in}\noindent
where $\sigma(\cdot)$ denotes the logistic function and the log-ratio score is
\vspace{-0.0in}\noindent
\begin{equation}
s_\theta(x,y)
= \beta \bigl(\log \pi_\theta(y\mid x) - \log \pi_{\mathrm{ref}}(y\mid x)\bigr),
\end{equation}
\vspace{-0.0in}\noindent
with inverse temperature $\beta>0$.

Sequence likelihoods are computed using summed token log-probabilities,
\vspace{-0.0in}\noindent
\begin{equation}\label{eq:sum_token_log_prob}
\log \pi_\theta(y\mid x)
= \sum_{t=1}^T \log \pi_\theta(y_t \mid x, y_{<t}),
\end{equation}
\vspace{-0.0in}\noindent
which implicitly discourages pathological verbosity without requiring explicit length penalties.

Given a preferred-dispreferred pair $(y^+,y^-)$, standard DPO minimizes
\vspace{-0.0in}\noindent
\begin{equation}
\mathcal{L}_{\mathrm{DPO}}(x)
= -\log \sigma\bigl(s_\theta(x,y^+) - s_\theta(x,y^-)\bigr).
\end{equation}
\vspace{-0.0in}\noindent
Although effective, this pairwise formulation ignores transitivity constraints when multiple rollouts are available, discarding global structure present in ranked samples. GraphDPO addresses this limitation by optimizing directly over preference graphs induced by partial orders.

\section{Graph Direct Preference Optimization}

We propose \emph{Graph Direct Preference Optimization} (GraphDPO), a generalization of DPO that leverages structured preference relations over multiple rollouts per prompt. Instead of treating preference comparisons as isolated pairs or strict lists, GraphDPO models them as a \emph{directed acyclic preference graph} and optimizes an objective defined over graph neighborhoods. In particular, GraphDPO constructs a preference graph over sampled responses for each prompt and aggregates supervision along directed edges. The resulting objective propagates the transitive structure of the input preferences and maintains linear computational complexity under the layered construction used in all our experiments.

\subsection{Preference Graph Representation}\label{sec:graph_repr}

Given a prompt $x$ and $K$ candidate responses $\{y_1,\ldots,y_K\}$ with preference relation $\succ_x$, we construct a directed acyclic graph (DAG)
\vspace{-0.0in}\noindent
\[
\mathcal{G}_x = (V,E),
\]
\vspace{-0.0in}\noindent
where $V = \{1,\ldots,K\}$ indexes responses and edges encode dominance relations. For responses $y_i$ and $y_j$,
\vspace{-0.0in}\noindent
\[
(i \rightarrow j) \in E
\quad \text{iff} \quad
y_i \succ_x y_j.
\]
\vspace{-0.0in}\noindent
Thus, each directed edge represents a preference constraint that $y_i$ should receive higher policy score than $y_j$. The resulting graph captures the full transitive structure implied by the preference relation.

It is convenient to represent the graph using an adjacency matrix
\vspace{-0.0in}\noindent
\[
A_{ij} =
\begin{cases}
1 & \text{if } y_i \succ_x y_j, \\
0 & \text{otherwise.}
\end{cases}
\]
\vspace{-0.0in}\noindent
Because $\succ_x$ is transitive and irreflexive, $A$ corresponds to a DAG.

\paragraph{Equivalence-class masking.}

In many alignment settings, preference signals are discrete (e.g., correctness indicators). Multiple responses may therefore share the same preference level. Using the equivalence-class representation introduced in Section~\ref{sec:preliminary}, we apply a masking rule that permits edges only across classes. Graph construction then follows a simple masking rule: edges are permitted only across classes. For $y_i \in \mathcal{C}_g$ and $y_j \in \mathcal{C}_h$,
\vspace{-0.0in}\noindent
\[
A_{ij} =
\begin{cases}
1 & g < h \\
0 & g = h .
\end{cases}
\]
\vspace{-0.0in}\noindent
This masking interpretation prevents edges between tied samples and ensures that intra-class comparisons contribute zero loss. As a result, GraphDPO avoids spurious gradients that arise from arbitrarily ordering nearly identical responses. Incomparable responses are handled by the same mechanism: two responses that cannot be compared (e.g., programs passing different subsets of unit tests at the same pass rate) are placed in the same class with no edge between them, and therefore exchange no gradient. All experiments in this paper use this construction, i.e., ranked equivalence classes with ties and incomparabilities.

\paragraph{General preference DAGs.}
The objective introduced below is defined over each node's dominated set and is therefore well-defined for any DAG, not only layered ones; the layered construction is what enables the linear-time implementation. More general preference structures are naturally expressible through edge weights $w_{ij}\in[0,1]$ that modulate each dominated response's contribution to the aggregation: noisy or uncertain comparisons receive fractional weights, missing comparisons correspond to $w_{ij}=0$ (absent edges), and confident comparisons to $w_{ij}=1$. This recovers the binary edge/no-edge construction used in our experiments as a special case. We do not evaluate weighted or non-layered graphs empirically and discuss them as future work in Section~\ref{sec:conclusion}.

\subsection{Graph-Structured Preference Objective}

Let $s_\theta(x,y)$ denote the DPO log-ratio score defined in Section~\ref{sec:preliminary}. For numerical stability, we apply per-prompt centering
\vspace{-0.0in}\noindent
\[
\tilde{s}_i
=
s_\theta(x,y_i)
-
\frac{1}{K}\sum_{j=1}^{K}s_\theta(x,y_j).
\]
\vspace{-0.0in}\noindent
This transformation preserves all pairwise score differences and leaves the objective invariant to per-prompt additive shifts, as $\tilde{s}$ appears only through differences and log-sum-exp terms.

Given the preference graph, we define the dominated neighborhood of node $i$ as
\vspace{-0.0in}\noindent
\[
\mathcal{N}^-(i)
=
\{ j \in V : A_{ij} = 1 \}.
\]
\vspace{-0.0in}\noindent
This set contains all responses that $y_i$ dominates according to the preference graph.

GraphDPO assigns likelihood to each node relative to its neighborhood using a Plackett--Luce choice model. Intuitively, response $y_i$ should be selected as the best element among the set consisting of itself and all dominated responses. The corresponding local loss is
\vspace{-0.0in}\noindent
\begin{equation}
\ell_i
=
-
\tilde{s}_i
+
\log
\sum_{k \in \{i\} \cup \mathcal{N}^-(i)}
\exp(\tilde{s}_k).
\end{equation}
\vspace{-0.0in}\noindent
The full graph loss aggregates contributions from all nodes with non-empty dominated neighborhoods:
\vspace{-0.0in}\noindent
\begin{equation}
\mathcal{L}_{\mathrm{graph}}(x)
=
\frac{1}{|\{i:\mathcal{N}^-(i)\neq\emptyset\}|}
\sum_{i:\mathcal{N}^-(i)\neq\emptyset}
\ell_i .
\end{equation}
\vspace{-0.0in}\noindent
This formulation encourages preference consistency across the entire graph: if $y_i$ dominates $y_j$, the objective encourages $s_i > s_j$, while the transitive relations present in the input propagate naturally through shared neighborhoods. We provide a formal discussion of the connection to Plackett--Luce models and the surrogate interpretation of our objective in Appendix~\ref{appd:connect_to_PL}, and analyze its optimization properties, including implicit constraint weighting and robustness, in Appendix~\ref{appd:opt_properties}.

\paragraph{Edge interpretation.}

Although the loss is expressed through node neighborhoods, it can equivalently be viewed as aggregating supervision over graph edges. Each directed edge $(i \rightarrow j)$ contributes a preference constraint encouraging
\vspace{-0.0in}\noindent
\[
s_\theta(x,y_i) > s_\theta(x,y_j),
\]
\vspace{-0.0in}\noindent
while the $\log\sum\exp$ aggregation couples these constraints across dominated responses. This highlights that GraphDPO optimizes preference structure encoded directly in the graph topology.

\paragraph{Computational Complexity.}
Each node contributes a single $\log\sum\exp$ over its dominated set. Under the equivalence-class (layered DAG) construction, dominated neighborhoods correspond to unions of lower-ranked groups, enabling efficient suffix aggregation and yielding $O(K)$ per-prompt complexity. For general partial orders without layered structure, the cost scales with the number of edges as $O(|E|)$, up to $O(K^2)$. In practice, many alignment settings with discrete preference signals naturally induce layered graphs, where the linear-time implementation applies. Appendix~\ref{appd:complexity} gives the explicit three-pass algorithm and a complexity lemma, and Appendix~\ref{appd:code} gives its PyTorch implementation.

\subsection{Ground-Truth Anchoring}

When a verified solution $y_{\mathrm{gt}}$ is available (e.g., the correct answer in reasoning benchmarks), we incorporate it as a dominant node in the preference graph. Let
\vspace{-0.0in}\noindent
\[
V_{\mathrm{worse}} =
\{ j \in V : y_{\mathrm{gt}} \succ_x y_j \}.
\]
\vspace{-0.0in}\noindent
The anchoring loss mirrors the graph objective:
\vspace{-0.0in}\noindent
\begin{equation}
\mathcal{L}_{\mathrm{anchor}}(x)
=
-
\tilde{s}_{\mathrm{gt}}
+
\log
\sum_{k \in \{\mathrm{gt}\} \cup V_{\mathrm{worse}}}
\exp(\tilde{s}_k).
\end{equation}
\vspace{-0.0in}\noindent
Ground-truth nodes participate only in this anchoring term and are excluded from $\mathcal{L}_{\mathrm{graph}}$, preventing conflicting gradients when generated responses match or exhibit equivalent preference signals.

\paragraph{Stabilization-to-relaxation schedule.}
We apply a time-dependent weight $\lambda_{\mathrm{gt}}(t)$ to the anchoring term. Early in training, the policy is weak and few rollouts are correct, so the graph provides limited positive signal; a relatively large $\lambda_{\mathrm{gt}}(t)$ stabilizes optimization by anchoring the policy to verified solutions. As training progresses, more rollouts achieve correctness and the graph itself supplies sufficient positive signal, so the weight is gradually reduced to unity, allowing greater flexibility among generated rollouts. Empirically, we find that initializing $\lambda_{\mathrm{gt}}$ to approximately $K/4$ to $K/3$, where $K$ is the number of rollouts per prompt, works well across tasks; this range arises from matching the gradient magnitude of the single anchor node to that of the graph term (Appendix~\ref{appd:lambda_sensitivity}), and performance is stable across a broad range of initialization values.

\subsection{Training Objective}\label{sec:method_objective}

To stabilize optimization and prevent excessive deviation from the reference policy, we include a token-level KL regularization term, commonly used in RLHF and DPO-style training. Specifically, for a response $y_i = (y_{i,1}, \ldots, y_{i,T_i})$, we compute
\begin{equation}
\mathcal{L}_{\mathrm{KL}}(x) =
\frac{1}{K} \sum_{i=1}^{K}
\frac{1}{T_i}
\sum_{t=1}^{T_i}
\Bigl(
\log \pi_\theta(y_{i,t} \mid x, y_{i,<t})
-
\log \pi_{\mathrm{ref}}(y_{i,t} \mid x, y_{i,<t})
\Bigr),
\end{equation}
corresponding to a sequence-mean-token-mean KL divergence between the policy and reference model. Because training is online, with rollouts $y_i$ sampled from the current policy $\pi_\theta$, this expectation is a Monte Carlo estimate of $\mathrm{KL}(\pi_\theta\,\|\,\pi_{\mathrm{ref}})\ge 0$, so minimizing it with a positive coefficient penalizes divergence from the reference policy. The KL coefficient $\lambda_{\mathrm{KL}}(t)$ follows a warmup--decay schedule that decays to zero over training, so the term stabilizes early optimization and vanishes by convergence. We treat this term purely as a regularizer; it does not affect the relative preference structure encoded by the graph objective.

\paragraph{Overall objective.}
The complete training objective for a prompt $x$ combines graph preference learning with optional ground-truth anchoring and KL regularization:
\begin{equation}
\mathcal{L}(x)
=
\mathcal{L}_{\mathrm{graph}}(x)
+
\lambda_{\mathrm{gt}}(t)\,\mathcal{L}_{\mathrm{anchor}}(x)
+
\lambda_{\mathrm{KL}}(t)\,\mathcal{L}_{\mathrm{KL}}(x),
\end{equation}
where $\lambda_{\mathrm{gt}}(t)$ follows the stabilization-to-relaxation schedule described above. When ground-truth solutions are unavailable, the anchoring term is omitted.

\paragraph{Reduction to DPO.}
GraphDPO recovers standard pairwise DPO as a special case. When $K=2$ with a strict preference $y_1 \succ y_2$, the preference graph contains a single edge $(1\rightarrow2)$. The graph loss reduces to
\vspace{-0.0in}\noindent
\[
-\tilde{s}_1 + \log(\exp(\tilde{s}_1)+\exp(\tilde{s}_2))
=
-\log \sigma(\tilde{s}_1-\tilde{s}_2),
\]
\vspace{-0.0in}\noindent
which is exactly the binary cross-entropy objective of DPO (up to centering).

\section{Experiments}

We assess GraphDPO across mathematical reasoning and program synthesis, spanning diverse task structures, difficulty levels, and base model configurations. For mathematical reasoning, we use GSM8K~\citep{cobbe2021training} and MATH~\citep{hendrycks2021measuring}, which range from elementary word problems to competition-level multi-step proofs, with \textbf{Qwen3-4B-Thinking-2507}, a reasoning-specialized model with explicit chain-of-thought capability. For program synthesis on the APPS benchmark~\citep{hendrycks2021apps}, we use \textbf{Qwen3-4B-Instruct-2507}, an instruction-tuned variant optimized for general task following. The use of two distinct base policies verifies that improvements are not policy-specific and that GraphDPO generalizes across alignment regimes. Additional experimental details are provided in Appendix~\ref{appd:exp_setting}, and a reference implementation of the loss in Appendix~\ref{appd:code}.

\paragraph{Baselines.}
We compare GraphDPO against representative alignment paradigms spanning supervised, pairwise, group-based, and listwise objectives. OTS denotes the off-the-shelf base model without additional alignment. SFT performs supervised fine-tuning on ground-truth trajectories. GRPO applies group-relative policy optimization via within-group reward normalization. SPIN is an iterative self-play method based on pairwise comparisons. MPO and SWEPO extend DPO to multi-sample settings via set-level and weighted groupwise objectives, respectively. PRO and LiPO are strict listwise objectives: PRO follows a Plackett--Luce formulation over total orders, while LiPO adopts a LambdaLoss-style ranking objective. Importantly, PRO serves as a structural ablation of GraphDPO that removes graph edge encoding and ground-truth anchoring, isolating the effect of graph-based factorization beyond standard listwise or multi-sample formulations.

\subsection{GSM8K: Mathematical Reasoning}

We evaluate on GSM8K using exact-match accuracy on the standard test split. 
All methods use Qwen3-4B-Thinking-2507 with identical rollout budgets and decoding configurations to ensure a controlled comparison.

\paragraph{Comparison to baselines.}
Results are shown in Table~\ref{tab:main_results}. We observe a steady improvement as stronger structural assumptions over rollouts are introduced. Moving from OTS to SFT yields +3.26 points, reflecting the benefit of supervised filtering. Pairwise methods (DPO, SPIN) provide further gains, while strict listwise objectives (PRO, LiPO) reach 88.82\% and 89.65\%, respectively. GraphDPO substantially outperforms all baselines, achieving 92.42\% without anchoring and 92.75\% with anchoring; the gap over the strongest baseline (SWEPO) is roughly 14$\times$ the pooled standard deviation. Because GSM8K correctness is binary, the realized preference graph is two-layer (correct vs.\ incorrect), and GraphDPO reduces to per-response dominated-set normalization with tie masking (Appendix~\ref{appd:opt_properties}); the gains here therefore come from this normalization rather than from graph depth. The improvement over listwise methods (PRO, LiPO) isolates the effect of equivalence-class edge encoding: while listwise methods impose strict total orders and propagate gradient through arbitrary orderings of tied samples, GraphDPO introduces no artificial comparisons among them. The comparison against DPO with the identical rollout budget separates two contributions: multi-sample structure (PRO over DPO, +1.77) and per-response graph factorization with tie masking (GraphDPO over PRO, +3.60).

\begin{table}[t]
\centering
\caption{GraphDPO performance across GSM8K, MATH-500, and APPS, with $K{=}8$ rollouts per prompt for all preference-based methods on all three benchmarks.
PRO corresponds to GraphDPO without graph edge encoding and ground-truth anchoring; DPO uses the top-1/bottom-1 pair from the same $K{=}8$ rollouts. All preference-based methods are reported as mean $\pm$ standard deviation over multiple seeds; OTS and SFT are single-run.}
\label{tab:main_results}
\setlength{\tabcolsep}{6pt}
\renewcommand{\arraystretch}{0.98}

\begin{tabular}{lccc}
\toprule
 & \multicolumn{3}{c}{\textbf{Acc. (\%)}} \\
\cmidrule(lr){2-4}
\textbf{Method} & \textbf{GSM8K} & \textbf{MATH-500} & \textbf{APPS} \\
\midrule
OTS   & 80.89 & 32.80 & 44.22 \\
SFT   & 84.15 & 54.80 & 59.08 \\
GRPO  & 85.59 {\scriptsize $\pm$ 0.30} & 84.00 {\scriptsize $\pm$ 0.28} & 68.67 {\scriptsize $\pm$ 1.18} \\
DPO   & 87.05 {\scriptsize $\pm$ 0.78} & 82.70 {\scriptsize $\pm$ 0.42} & 68.35 {\scriptsize $\pm$ 1.06} \\
SPIN  & 88.10 {\scriptsize $\pm$ 0.86} & 79.80 {\scriptsize $\pm$ 0.57} & 65.20 {\scriptsize $\pm$ 1.13} \\
PRO   & 88.82 {\scriptsize $\pm$ 0.16} & 82.30 {\scriptsize $\pm$ 1.84} & 68.84 {\scriptsize $\pm$ 0.37} \\
LiPO  & 89.65 {\scriptsize $\pm$ 0.59} & 85.40 {\scriptsize $\pm$ 0.28} & 68.96 {\scriptsize $\pm$ 0.51} \\
MPO   & 89.98 {\scriptsize $\pm$ 0.35} & 84.50 {\scriptsize $\pm$ 0.42} & 67.72 {\scriptsize $\pm$ 0.73} \\
SWEPO & 91.29 {\scriptsize $\pm$ 0.11} & 85.00 {\scriptsize $\pm$ 0.85} & 67.38 {\scriptsize $\pm$ 1.44} \\
\midrule
GraphDPO (w/o GT)
& 92.42 {\scriptsize $\pm$ 0.33}
& 87.40 {\scriptsize $\pm$ 0.20}
& 72.93 {\scriptsize $\pm$ 0.72} \\

\textbf{GraphDPO (w/ GT)}
& \textbf{92.75} {\scriptsize $\pm$ 0.09}
& \textbf{88.87} {\scriptsize $\pm$ 0.61}
& \textbf{73.76} {\scriptsize $\pm$ 0.19} \\
\bottomrule
\end{tabular}
\end{table}

\begin{wraptable}{r}{0.5\textwidth}\vspace{-0.26in}
% \begin{table}[t]
\centering
\caption{Ablation study of GraphDPO on GSM8K with varying group size $K$ 
and with/without ground-truth anchoring.}
\label{tab:gsm8k_ablation}
\setlength{\tabcolsep}{6pt}
\renewcommand{\arraystretch}{0.98}

\begin{tabular}{lcccc}
\toprule
 & \textbf{$K=2$} & \textbf{$K=4$} & \textbf{$K=6$} & \textbf{$K=8$} \\
\midrule
w/o GT & 88.78 & 92.04 & 92.13 & 92.42 \\
w/ GT  & 91.81 & 92.12 & 92.65 & \textbf{92.75} \\
\bottomrule
\end{tabular}\vspace{-0.15in}
% \end{table}
\end{wraptable}

\paragraph{Ablation study.}
Besides the comparison to PRO (Table~\ref{tab:main_results}) that isolates the effect of graph-structured supervision from standard multi-sample (listwise) aggregation, Table~\ref{tab:gsm8k_ablation} studies the effect of ground-truth anchoring and rollout count $K$, which controls the density of the induced preference graph. Firstly, anchoring is most valuable at small rollout budgets: the gain is largest at $K=2$ (+3.03 points), where the graph reduces to a single pairwise comparison and the anchor supplies a reliable positive node, and shrinks to +0.33 at $K=8$, comparable to seed-level noise. We therefore characterize anchoring as a low-$K$ stabilizer whose benefit a richer graph largely subsumes. Secondly, increasing $K$ strengthens performance by enabling richer relational structure, with gains saturating between $K=4$ and $K=8$ (+0.38 without anchoring); we use this saturation on GSM8K to fix $K=8$ for all benchmarks, isolating the effect of the objective from the rollout budget. The benefit of the graph objective is not an artifact of rollout count: a baseline $K$-scaling comparison (Appendix~\ref{appd:kscaling}) shows that PRO with the same $K$ cannot close the gap at $K=2$, $4$, or $8$, and adding the same gold trajectory to PRO yields no gain, confirming that GraphDPO's advantage comes from the graph-structured objective rather than the anchor.

Overall, the GSM8K results support three claims:
(i) respecting ties through equivalence-class masking improves over strict listwise objectives;
(ii) aggregating supervision over dominated neighborhoods yields stronger alignment than pairwise reductions at the same rollout budget; and
(iii) ground-truth anchoring provides additional stability under limited rollout budgets.

% \vspace{-0.1in}
\subsection{MATH-500: Advanced Reasoning}
% \vspace{-0.1in}

We evaluate on MATH-500~\citep{lightman2023lets} using exact-match accuracy, focusing on generalization from GSM8K-style supervision to substantially harder multi-step problems. While the supervision signal remains binary, the distribution of rollouts becomes significantly more diverse due to longer reasoning chains and more complex failure modes.

\paragraph{Comparison to baselines.}
As shown in Table~\ref{tab:main_results}, GraphDPO achieves 87.40\% without anchoring and 88.87\% with anchoring, outperforming all baselines by a clear margin (+3.47 over the strongest baseline, LiPO at 85.40\%). Notably, the gap between GraphDPO and strong listwise methods (LiPO: 85.40\%, PRO: 82.30\%) is larger than on GSM8K, suggesting that the benefits of graph-structured preference modeling become more pronounced as task complexity increases. A key observation is that several baselines exhibit inconsistent scaling behavior when moving from GSM8K to MATH-500. For example, SPIN underperforms relative to both GRPO and listwise methods, indicating that iterative pairwise updates may struggle to propagate consistent signals over long reasoning trajectories. In contrast, GRPO and LiPO remain competitive, but still fall short of GraphDPO, suggesting that while group-level normalization and total-order modeling help, they do not fully capture the structure of complex reasoning outcomes.

\paragraph{Effect of structured supervision under harder tasks.}
The stronger gains on MATH-500 highlight an important distinction: as reasoning depth increases, errors become more heterogeneous, and rollouts often differ substantially in reasoning trajectories and intermediate steps, even when final correctness is binary. This makes strict total-order assumptions more brittle: forcing diverse correct (or diverse incorrect) trajectories into an arbitrary ranking injects gradient noise precisely where trajectories differ most. GraphDPO avoids this by placing tied responses in one equivalence class with no edges between them, so multiple distinct correct trajectories coexist without forced ordering decisions. This leads to more effective credit assignment: instead of pushing the model toward a single ``best'' trajectory, GraphDPO aggregates supervision over neighborhoods of dominated rollouts, reinforcing consistent improvements across multiple reasoning paths. The result is a more stable optimization signal that better aligns with the underlying structure of difficult problems.

\paragraph{Role of anchoring.}
Ground-truth anchoring provides a +1.47 point gain on MATH-500, which is larger than on GSM8K and comfortably exceeds the multi-seed standard deviation of both configurations. This suggests that as task difficulty increases, having a reliable global reference becomes more important for stabilizing preference propagation. In complex reasoning settings where most sampled rollouts are imperfect, anchoring helps prevent drift by ensuring that the preference graph remains calibrated to correct solutions. Overall, the MATH-500 results demonstrate that GraphDPO is particularly effective in regimes where reasoning is deep, errors are diverse, and simple ordering assumptions break down. The advantage over both pairwise and listwise baselines indicates that explicitly modeling structured relationships among rollouts is critical for scaling alignment to harder problem domains.

% \vspace{-0.1in}
\subsection{APPS: Program Synthesis}
% \vspace{-0.1in}

We further evaluate on the APPS benchmark, which measures program synthesis via pass rate on hidden unit tests. In contrast to GSM8K's exact answer matching, APPS provides execution-based supervision over full programs, introducing longer structured outputs and partial correctness signals. We adopt an instruction-tuned base policy to test whether the benefits of GraphDPO extend beyond reasoning-specialized initialization.

The results are reported in Table~\ref{tab:main_results}, where GraphDPO achieves 72.93\% without anchoring and 73.76\% with anchoring, outperforming all baselines. Notably, it improves over both group-based (GRPO) and strict listwise methods (PRO, LiPO), indicating that modeling preference structure as a graph provides a stronger inductive bias than normalization-based or total-order objectives. The improvement over GRPO is particularly notable given that it already leverages group-level normalization; the additional gains indicate that explicitly modeling the dominance relations between preference levels provides stronger constraints than averaging-based comparisons.

Unlike GSM8K and MATH, APPS evaluates programs through hidden unit tests, where feedback reflects partial correctness across test cases. Equivalence classes are defined by the fraction of hidden tests passed (Appendix~\ref{appd:exp_setting}), so APPS is the benchmark where the preference graph is genuinely multi-layer: all prompts admit at least 3 achievable preference levels (median 11), and 17.8\% of prompts realize 3 or more layers from base-model rollouts at $K{=}8$ (Appendix~\ref{appd:layer_stats}). In this setting, scalar reward normalization or strict total-order assumptions (as in PRO and LiPO) can blur fine-grained relational signals between rollouts that pass different subsets of tests. By constructing a preference graph over equivalence classes, GraphDPO preserves these distinctions and aggregates supervision over dominated neighborhoods. Consistent with this, APPS is where GraphDPO's margin over the strongest baseline is largest (+4.80 over LiPO), and a depth-stratified analysis (Appendix~\ref{appd:layer_stats}) shows the improvement over the base policy growing monotonically with realized layer count (+16.8 at 2 layers, +25.9 at 3, +36.2 at $\ge$4). The persistence of gains under an instruction-tuned base model indicates that the improvement is not tied to reasoning-specialized initialization, but arises from the structural inductive bias of graph-based preference modeling.

% \vspace{-0.1in}
\section{Conclusion}\label{sec:conclusion}
% \vspace{-0.1in}

We present GraphDPO, a graph-structured generalization of DPO for aligning language models from preference relations over multiple rollouts. GraphDPO represents comparisons as a directed acyclic preference graph, capturing ties and incomparability, propagating the transitive structure of the input preferences, and avoiding spurious supervision through equivalence-class edge masking. The resulting objective aggregates supervision over dominated neighborhoods while maintaining linear computational complexity. Experiments show that GraphDPO outperforms supervised, pairwise, and listwise baselines across reasoning and program synthesis tasks. Ablations further highlight the impact of richer preference graphs and optional ground-truth anchoring on stabilizing early optimization and improving final performance, suggesting that modeling rollout preferences as graphs provides a simple and effective inductive bias for LLM alignment. Extending graph-structured preference optimization to long-horizon agentic trajectories, such as multi-step procedures executed by LLM agents~\citep{liu2026eluna}, is a natural next step.

\paragraph{Broader Impacts.}
Improving preference optimization can lead to more reliable and controllable language models. However, if preference signals are biased or manipulated, the method may reinforce undesirable behaviors, highlighting the need for careful data curation.

\paragraph{Limitations.}
GraphDPO relies on the quality of sampled rollouts, and may amplify biases present in the underlying preference signals. Our experiments cover layered constructions (ranked equivalence classes with ties and incomparabilities) induced by discrete verification signals; non-layered partial orders and noisy human preferences, which the framework expresses through edge weights (Section~\ref{sec:graph_repr}), are not evaluated empirically and remain future work. Our oracle trajectories wrap reference solutions in the model's \texttt{<think>} format, which is distributionally close to the policy prior and may inflate the apparent anchoring benefit relative to settings where the oracle is genuinely out-of-distribution. All experiments use Qwen3 (4B Thinking and Instruct) base models, and the connection to Plackett--Luce is a surrogate objective in the general case, with exact equivalence holding only for strict total orders or layered graphs with layer-wise independence (Appendix~\ref{appd:connect_to_PL}). Finally, while the method is computationally efficient in the number of edges, scaling to very large candidate sets can still introduce overhead.

% \begin{ack}
% Use unnumbered first level headings for the acknowledgments. All acknowledgments

% Do {\bf not} include this section in the anonymized submission, only in the final paper. You can use the \texttt{ack} environment provided in the style file to automatically hide this section in the anonymized submission.
% \end{ack}

%\section*{References}
% \bibliographystyle{plainnat}
\bibliographystyle{unsrtnat}

%%%%%%%%%%%%%%%%%%%%%%%%%%%%%%%%%%%%%%%%%%%%%%%%%%%%%%%%%%%%
\newpage
\appendix

\section{Experimental Settings}\label{appd:exp_setting}

\paragraph{Optimization.}
Across all experiments, we use AdamW with $(\beta_1, \beta_2) = (0.9, 0.999)$ and weight decay $10^{-2}$. The learning rate follows a cosine schedule with warmup, decaying to $10\%$ of the peak value. Training proceeds until convergence. All methods share the same optimizer and schedule for controlled comparison.

\paragraph{Infrastructure.}
All experiments are conducted on a single node with 8 NVIDIA H100 GPUs, each with 80GB memory. Our implementation is built on the verl framework~\citep{sheng2024hybridflow} (Appendix~\ref{appd:code}). We use FSDP2 for model sharding and vLLM for rollout generation with tensor parallelism of 1.

\subsection{GSM8K}

\paragraph{Dataset and setup.}
We use the GSM8K~\citep{cobbe2021training} training split (7{,}473 examples) for training and evaluate on the standard test split (1{,}319 examples) using exact-match accuracy. The base model is \textbf{Qwen3-4B-Thinking-2507}, a reasoning-specialized model with explicit chain-of-thought capability.

\paragraph{Ground-truth trajectory generation.}
For ground-truth anchoring, we construct oracle trajectories by wrapping the reasoning portion of each GSM8K reference answer with the model's thinking tokens (\texttt{<think>...</think>}), producing trajectories compatible with the model's chain-of-thought format.

\paragraph{Training configuration.}
We sample $K$ rollouts per prompt with temperature $0.8$. The maximum prompt and response lengths are both set to 1{,}024 tokens. We use a training batch size of 32 prompts per step. The inverse temperature is set to $\beta{=}0.05$. For ground-truth anchoring, the weight $\lambda_{\mathrm{gt}}(t)$ is linearly decayed from $2.5$ to $1.0$ over the course of training. The reference policy is held fixed throughout training (i.e., no periodic reference updates as in SPIN).

\subsection{MATH}

\paragraph{Dataset and filtering.}
We evaluate on the MATH benchmark~\cite{hendrycks2021measuring} using exact-match accuracy. From the full training set (7{,}500 problems across five difficulty levels), we construct a training subset of 1{,}000 problems by subsampling each difficulty level at twice the proportion present in the test split, yielding a level distribution that mirrors the evaluation setting (Level~1: 86, Level~2: 180, Level~3: 210, Level~4: 256, Level~5: 268). We evaluate on MATH-500~\cite{lightman2023lets}, a representative 500-problem subset of the standard test split. This subsampling ensures that training supervision is balanced across difficulty levels rather than dominated by easier problems.

\paragraph{Ground-truth trajectory generation.}
For ground-truth anchoring, we construct oracle trajectories by wrapping the reference solution of each MATH problem with the model's thinking tokens (\texttt{<think>...</think>}), producing trajectories compatible with the model's chain-of-thought format.

\paragraph{Training configuration.}
The base model is \textbf{Qwen3-4B-Thinking-2507}, the same reasoning-specialized model used for GSM8K. We sample $K{=}8$ rollouts per prompt with temperature $0.8$. The maximum prompt length is set to 3{,}000 tokens and the maximum response length to 2{,}000 tokens to accommodate multi-step reasoning chains. We use a training batch size of 32 prompts per step with a peak learning rate of $2 \times 10^{-6}$. The inverse temperature is set to $\beta{=}0.002$.

\subsection{APPS}

\paragraph{Dataset and filtering.}
We use the APPS benchmark~\cite{hendrycks2021apps} for program synthesis evaluation. From the original training split (5{,}000 problems), we apply the following filtering criteria: (i)~problems requiring starter code are removed, (ii)~problems with function-based input/output specifications (i.e., containing a \texttt{fn\_name} key) are excluded, retaining only stdin/stdout-style problems, (iii)~problems with no test case inputs are discarded, and (iv)~example input/output pairs embedded in problem descriptions are removed, as we find these are identical to the hidden unit tests used for evaluation, which would otherwise enable reward hacking. After filtering, 1{,}644 training problems remain. For evaluation, we apply the same filtering to the test split and randomly subsample 1{,}000 problems. Performance is measured by execution-based pass rate against hidden unit tests.

\paragraph{Evaluation protocol.}
Our APPS evaluation uses a filtered subset designed to remove data leakage and ensure reliable execution-based evaluation. While this departs from standard splits, we apply identical preprocessing across all methods. We acknowledge that absolute performance numbers are not directly comparable to prior work and focus on relative improvements under controlled settings.

\paragraph{Equivalence-class construction.}
On APPS, equivalence classes are defined by the fraction of hidden tests passed (up to 10 test groups per problem, yielding up to 11 distinct pass-rate levels). Responses that achieve the same pass rate are placed in the same class: this includes both exact ties and responses that pass \emph{different} subsets of tests at the same pass rate, which are incomparable and receive no edge (zero gradient). Classes are ordered by pass rate, so the construction is layered and the $O(K)$ implementation of Appendix~\ref{appd:complexity} applies directly.

\paragraph{Ground-truth trajectory generation.}
For the anchored variant, we use the human-written reference solutions provided in the APPS dataset as oracle trajectories.

\paragraph{Training configuration.}
The base model is \textbf{Qwen3-4B-Instruct-2507}, an instruction-tuned variant. We sample $K{=}8$ rollouts per prompt with temperature $0.8$. The maximum prompt length is set to 2{,}048 tokens and the maximum response length to 7{,}000 tokens to accommodate full program outputs. We use a training batch size of 32 prompts per step with a peak learning rate of $2 \times 10^{-6}$. The inverse temperature is set to $\beta{=}0.5$. The reference policy is held fixed throughout training.

\section{Linear-Time Algorithm and Complexity Analysis}
\label{appd:complexity}

We give the explicit algorithm underlying the $O(K)$ complexity claim and state the conditions under which it holds.

\paragraph{Setting.}
Let $K$ responses be bucketed into ordered equivalence classes $\mathcal{C}_1 \succ \cdots \succ \mathcal{C}_G$ (the layered construction of Section~\ref{sec:graph_repr}), with centered scores $\tilde{s}_1,\ldots,\tilde{s}_K$. Under this construction, the dominated neighborhood of any node $i \in \mathcal{C}_g$ is the union of all lower-ranked classes, $\mathcal{N}^-(i) = \bigcup_{g'>g} \mathcal{C}_{g'}$, so all nodes in the same class share the same dominated set.

\paragraph{Algorithm.}
The full graph loss is computed in three passes:
\begin{enumerate}
\item \textbf{Bucketing} ($O(K)$): assign each response to its class and compute class sums $S_g = \sum_{k\in\mathcal{C}_g}\exp(\tilde s_k)$ for $g = 1,\ldots,G$.
\item \textbf{Backward scan} ($O(G)$): compute suffix sums $Z_g = S_g + Z_{g+1}$ for $g = G, G{-}1, \ldots, 1$, with $Z_{G+1} = 0$. The quantity $Z_{g+1}$ equals $\sum_{k \in \mathcal{N}^-(i)} \exp(\tilde{s}_k)$ for every $i \in \mathcal{C}_g$.
\item \textbf{Per-node loss} ($O(1)$ each, $O(K)$ total): for each $i\in\mathcal{C}_g$ with $g < G$, compute $\ell_i = -\tilde s_i + \log\bigl(\exp(\tilde s_i) + Z_{g+1}\bigr)$.
\end{enumerate}

\paragraph{Complexity lemma.}
\emph{Under the layered construction, the total cost of computing $\mathcal{L}_{\mathrm{graph}}$ is $O(K)$, for any number of classes $G \le K$.}

\emph{Proof.} Pass 1 touches each of the $K$ responses once. Pass 2 performs $G \le K$ additions. Pass 3 computes one $O(1)$ expression per node using the precomputed suffix sum, totaling $O(K)$. Summing, the cost is $O(K) + O(G) + O(K) = O(K)$. \hfill $\square$

\paragraph{Scope of the bound.}
The $O(K)$ bound relies on all nodes within a class sharing the same dominated set, which is exactly the layered structure. For a general partial order, dominated sets need not be nested or shared, and the cost of the objective scales with the number of edges as $O(|E|)$, which is $O(K^2)$ in the worst case. All experiments in this paper use the layered construction, where the linear bound applies.

\section{Reference Implementation}
\label{appd:code}

Our implementation is built on the open-source verl framework~\citep{sheng2024hybridflow}, starting from its SPIN recipe and replacing the pairwise chosen/rejected DPO update with the GraphDPO objective. Rollout generation (vLLM), FSDP2 sharding, and token-level log-probability computation use verl's standard components unchanged. Our modifications are: (i)~for each prompt, all $K$ rollouts are kept as a group rather than reduced to a chosen/rejected pair; (ii)~each response is assigned a scalar preference score by the task verifier (exact-match correctness on GSM8K and MATH, fraction of hidden tests passed on APPS), and responses with equal preference scores form an equivalence class; (iii)~when ground-truth anchoring is enabled, the oracle trajectory is appended to the group as an additional node, scored by the same verifier, and assigned the per-sample weight $\lambda_{\mathrm{gt}}(t)$, while all generated responses receive weight $1$; (iv)~the reference policy is held fixed; and (v)~the GraphDPO loss of Listing~\ref{lst:graphdpo} is combined with the KL term of Section~\ref{sec:method_objective}. Listing~\ref{lst:graphdpo} gives the loss used in all experiments: each node normalizes over itself and all strictly lower-ranked classes. Its explicit loop costs $O(K^2)$ per prompt, which is negligible relative to the policy forward passes at $K \le 9$. Listing~\ref{lst:graphdpo_linear} gives an equivalent vectorized implementation of the three-pass procedure of Appendix~\ref{appd:complexity}. Because preference signals in our settings are discrete (correctness on GSM8K and MATH, pass-rate level on APPS), each response is mapped to an integer preference level in $\{0,\ldots,L{-}1\}$, so bucketing is a direct index operation with no sorting or hashing, and the total cost is $O(K+L)$ per prompt, i.e., $O(K)$ for $L \le K$. The two listings agree to floating-point precision in both loss and gradients.

\begin{lstlisting}[style=pystyle, caption={GraphDPO loss used in all experiments (PyTorch).}, label={lst:graphdpo}]
def graphdpo_loss(policy_logps, ref_logps, scores, beta, sample_weights=None):
    # policy_logps, ref_logps: (B, K) summed token log-probs per response
    # scores: (B, K) preference scores; equal scores share a class
    # sample_weights: (B, K) lambda_gt(t) on the anchor node, 1 elsewhere
    s = beta * (policy_logps - ref_logps)
    s = s - s.mean(dim=1, keepdim=True)          # per-prompt centering
    B, K = s.shape
    prompt_losses = []
    for b in range(B):
        # class rank (0 = best class)
        _, inv = torch.unique(scores[b], sorted=True, return_inverse=True)
        rank = inv.max() - inv
        w = sample_weights[b] if sample_weights is not None else torch.ones_like(s[b])
        losses, weights = [], []
        for i in range(K):
            dominated = torch.where(rank > rank[i])[0]       # N^-(i)
            if dominated.numel() == 0:
                continue
            logits = torch.cat([s[b, i:i+1], s[b, dominated]])
            losses.append(torch.logsumexp(logits, dim=0) - s[b, i])  # ell_i
            weights.append(w[i])
        if losses:
            l, wt = torch.stack(losses), torch.stack(weights)
            prompt_losses.append((wt * l).sum() / wt.sum().clamp(min=1e-8))
    return torch.stack(prompt_losses).mean()
\end{lstlisting}

\begin{lstlisting}[style=pystyle, caption={Equivalent linear-time GraphDPO loss (three-pass procedure of Appendix~\ref{appd:complexity}).}, label={lst:graphdpo_linear}]
def graphdpo_loss_linear(policy_logps, ref_logps, levels, num_levels, beta,
                         sample_weights=None):
    # levels: (B, K) integer preference levels in [0, num_levels), higher = preferred
    s = beta * (policy_logps - ref_logps)
    s = s - s.mean(dim=1, keepdim=True)
    B, L = s.shape[0], num_levels
    w = sample_weights if sample_weights is not None else torch.ones_like(s)
    ninf = s.new_full((B, L), -torch.inf)
    # Pass 1: bucketing -- per-class log-sum-exp S_g (max-shifted for stability)
    counts = levels.new_zeros(B, L).scatter_add_(1, levels, torch.ones_like(levels))
    m = ninf.scatter_reduce(1, levels, s.detach(), "amax")
    m = torch.where(counts > 0, m, torch.zeros_like(m))
    sums = s.new_zeros(B, L).scatter_add(1, levels, torch.exp(s - m.gather(1, levels)))
    S = torch.where(counts > 0, m + torch.log(sums.clamp(min=1e-30)), ninf)
    # Pass 2: scan -- log-sum-exp over all strictly lower classes
    C = torch.logcumsumexp(S, dim=1)
    lower = torch.cat([ninf[:, :1], C[:, :-1]], dim=1)
    n_lower = counts.cumsum(1) - counts                   # |N^-(i)| per class
    # Pass 3: per-node loss ell_i
    ell = torch.logaddexp(s, lower.gather(1, levels)) - s
    mask = (n_lower.gather(1, levels) > 0).float()
    num, den = (w * mask * ell).sum(1), (w * mask).sum(1)
    valid = mask.sum(1) > 0
    return (num[valid] / den[valid].clamp(min=1e-8)).mean()
\end{lstlisting}

\section{Connection to Plackett--Luce and Surrogate Interpretation}
\label{appd:connect_to_PL}

\paragraph{Overview.}
GraphDPO is inspired by the Plackett--Luce (PL) family of choice models, but does not define a globally normalized likelihood over permutations or partial orders. Instead, it should be understood as a structured surrogate objective that decomposes preference learning into a collection of local choice problems defined over a preference graph.

\paragraph{Local multinomial choice interpretation.}
Recall that each term in the GraphDPO objective is
\[
\ell_i = -s_i + \log \sum_{k \in \{i\} \cup \mathcal{N}^-(i)} \exp(s_k),
\]
which corresponds to the negative log-probability of selecting response $y_i$ from the set $\{i\} \cup \mathcal{N}^-(i)$ under a multinomial logit (MNL) model:
\[
P(i \mid \{i\} \cup \mathcal{N}^-(i)) = 
\frac{\exp(s_i)}{\sum_{k \in \{i\} \cup \mathcal{N}^-(i)} \exp(s_k)}.
\]
Thus, each node defines a local choice problem in which $y_i$ is preferred over all responses it dominates.

\paragraph{Relation to Plackett--Luce models.}
The standard PL model defines a probability distribution over permutations via a sequential elimination process, where items are selected one at a time from a shrinking candidate set. GraphDPO can be viewed as a relaxation of this process: instead of imposing a single global elimination order, it defines multiple local choice problems over overlapping subsets induced by the preference graph.

When the preference graph corresponds to a strict total order (i.e., a chain), the dominated neighborhoods recover nested suffix sets, and the GraphDPO objective resembles a partial factorization of the PL likelihood. However, for general partial orders, the local choice sets overlap and do not correspond to a valid factorization of a globally normalized distribution over permutations.

\paragraph{Layer-wise elimination perspective.}
An alternative interpretation connects GraphDPO to a stagewise elimination process. Under the equivalence-class construction, responses are partitioned into ordered groups. One can view training as selecting preferred responses from progressively expanding candidate sets that include all dominated groups. This yields a sequence of local choice problems analogous to partial elimination in PL models, but without committing to a single global ordering. GraphDPO can thus be interpreted as approximating a layered PL factorization, where elimination is defined over groups rather than individual permutations.

\paragraph{Surrogate objective and consistency.}
Although GraphDPO does not define a globally normalized likelihood, it can be interpreted as a structured surrogate objective for the pairwise dominance constraints. For every edge $(i \to j)$ in the preference graph, $j \in \mathcal{N}^-(i)$, and the loss $\ell_i$ encourages $s_i > s_j$ through the shared normalization term. Aggregating over all nodes yields a coupled objective that encourages all pairwise constraints while promoting consistency with the transitive relations in the input. We do not claim a statistical consistency or calibration result (in the sense of, e.g., ListMLE for total orders) for the general overlapping-set case; Proposition~1 in Appendix~\ref{appd:opt_properties} characterizes the unregularized optimum.

Importantly, the use of shared log-sum-exp normalization across dominated sets couples multiple comparisons without explicitly enumerating $O(K^2)$ pairs. While individual edges may appear in multiple neighborhoods, this does not lead to inconsistent supervision; rather, it reinforces transitive relations through multiple overlapping local contexts.

\paragraph{Connection to pairwise DPO.}
When $K=2$, the graph reduces to a single edge, and the objective recovers standard DPO:
\[
\ell = -\log \sigma(s_1 - s_2).
\]
More generally, GraphDPO can be viewed as a structured generalization of pairwise DPO in which multiple comparisons are jointly normalized within dominated sets. This perspective places GraphDPO between pairwise and listwise approaches. Compared to pairwise methods, it captures higher-order structure by coupling comparisons through shared normalization. Compared to listwise PL objectives, it avoids requiring strict total orders or permutation-level normalization, making it more suitable for partial orders with ties. This trade-off yields a tractable objective that preserves transitive structure while remaining computationally efficient.

\paragraph{Conditions for equivalence to Plackett--Luce likelihood.}
The GraphDPO objective can be interpreted as a valid Plackett--Luce likelihood under a restricted set of assumptions. In particular, if the preference graph forms a strict total order (i.e., a chain), then the dominated neighborhoods define a nested sequence of candidate sets, and the product of local choice probabilities recovers a factorization of the standard PL likelihood. More generally, if the graph admits a layered structure and one assumes independence across layer-wise elimination steps, the objective corresponds to a group-wise PL model where selection is performed over equivalence classes rather than individual items. However, for arbitrary partial orders with overlapping dominated neighborhoods, these independence assumptions no longer hold, and the product of local choice terms does not correspond to a globally normalized likelihood. In this case, GraphDPO should be understood as a structured surrogate objective that approximates PL-style elimination while remaining computationally tractable.

\section{Optimization Properties and Implicit Weighting}
\label{appd:opt_properties}

\paragraph{Overview.}
While Appendix~\ref{appd:connect_to_PL} provides an interpretation of GraphDPO as a structured surrogate objective, here we analyze its optimization properties. In particular, we further characterize (i) pairwise consistency from an optimization perspective, (ii) implicit edge weighting induced by node-wise aggregation, and (iii) robustness under noisy or imbalanced preference signals.

\paragraph{Pairwise consistency.}
We first formalize the relationship between the GraphDPO objective and pairwise preference constraints.

\textbf{Proposition 1 (Pairwise consistency).}
\emph{
Let $G = (V, E)$ be a preference DAG. If $\mathcal{L}_{\mathrm{graph}} = \sum_{i \in V} \ell_i$ is minimized such that $\ell_i \to 0$ for all $i$, then the resulting scores satisfy $s_i \ge s_j$ for all $(i \to j) \in E$ up to arbitrarily small slack.
}

\emph{Proof.}
Fix any node $i$ and write $p_i(i) = \exp(s_i)/\sum_{k \in \{i\} \cup \mathcal{N}^-(i)} \exp(s_k)$, so that $\ell_i = -\log p_i(i)$. Suppose $\ell_i \le \varepsilon$ for some $\varepsilon > 0$; then $p_i(i) \ge e^{-\varepsilon}$. For any $j \in \mathcal{N}^-(i)$, since the softmax probabilities over $\{i\} \cup \mathcal{N}^-(i)$ sum to one,
\[
p_i(j) \;=\; \frac{\exp(s_j)}{\sum_{k \in \{i\} \cup \mathcal{N}^-(i)} \exp(s_k)} \;\le\; 1 - p_i(i) \;\le\; 1 - e^{-\varepsilon}.
\]
Dividing the two bounds,
\[
\exp(s_j - s_i) \;=\; \frac{p_i(j)}{p_i(i)} \;\le\; \frac{1 - e^{-\varepsilon}}{e^{-\varepsilon}} \;=\; e^{\varepsilon} - 1,
\]
hence $s_j \le s_i + \log(e^{\varepsilon} - 1)$. For $\varepsilon < \log 2$ the slack term $\log(e^{\varepsilon}-1)$ is negative, so $s_i > s_j$ strictly, and as $\varepsilon \to 0$ the required score gap $s_i - s_j \ge -\log(e^{\varepsilon}-1)$ grows without bound. Since $(i \to j) \in E$ implies $j \in \mathcal{N}^-(i)$, applying the argument to every node enforces all edge constraints. \hfill $\square$

\emph{Remark (role of regularization).}
Driving $\ell_i \to 0$ requires unbounded score gaps, whereas the KL term keeps scores finite. Proposition~1 therefore characterizes the optimum of the unregularized objective; because $\lambda_{\mathrm{KL}}(t)$ decays to zero over training, the trained model approaches this regime and satisfies a soft version of the constraints, with score gaps bounded by the residual per-node losses via the displayed inequality.

\paragraph{Implicit edge weighting.}
Unlike pairwise objectives that assign uniform weight to each edge, GraphDPO aggregates node-wise losses:
\[
\mathcal{L}_{\mathrm{graph}} = \sum_{i \in V} \ell_i.
\]
This induces a non-uniform weighting over edges. Specifically, an edge $(i \to j)$ contributes:
\begin{itemize}
\item directly through $\ell_i$,
\item indirectly through all $\ell_k$ such that $j \in \mathcal{N}^-(k)$.
\end{itemize}

\textbf{Proposition 2 (Degree-weighted constraints).}
\emph{
The effective weight associated with a node $j$ (and edges incident to it) scales with the number of ancestors of $j$ in the preference graph, i.e., the number of nodes whose dominated neighborhoods include $j$.
}

\emph{Proof.}
Let $\mathcal{A}(j) = \{i \in V : j \in \mathcal{N}^-(i)\}$ denote the ancestors of $j$. Only the losses $\ell_i$ with $i \in \mathcal{A}(j)$ depend on $s_j$, and each contributes gradient $\partial \ell_i / \partial s_j = p_i(j) \in (0,1)$, where $p_i(j)$ is the softmax probability of $j$ in the local choice set of $i$ (derived below). Hence
\[
\frac{\partial \mathcal{L}_{\mathrm{graph}}}{\partial s_j} = \sum_{i \in \mathcal{A}(j)} p_i(j),
\]
a sum of $|\mathcal{A}(j)|$ strictly positive terms. In particular, the aggregate gradient is upper-bounded by $|\mathcal{A}(j)|$ and, whenever the softmax weights are bounded below (e.g., early in training when scores are not yet separated, $p_i(j) \approx 1/|\{i\}\cup\mathcal{N}^-(i)| \ge 1/K$), it is lower-bounded by $|\mathcal{A}(j)|/K$, so the effective weight scales with the ancestor count. \hfill $\square$

\paragraph{Gradient-based characterization.}
We make the notion of implicit weighting precise by examining gradients. For any node $i$, the loss is
\[
\ell_i = -s_i + \log \sum_{k \in \{i\} \cup \mathcal{N}^-(i)} \exp(s_k).
\]
The gradient with respect to a dominated node $j \in \mathcal{N}^-(i)$ is
\[
\frac{\partial \ell_i}{\partial s_j}
=
\frac{\exp(s_j)}{\sum_{k \in \{i\} \cup \mathcal{N}^-(i)} \exp(s_k)}
\;\coloneqq\;
p_i(j),
\]
where $p_i(j)$ denotes the softmax probability of $j$ within the local choice set of $i$.

Aggregating over all nodes, the total gradient contribution on $s_j$ is
\[
\frac{\partial \mathcal{L}_{\mathrm{graph}}}{\partial s_j}
=
\sum_{\substack{i \in V \\ j \in \mathcal{N}^-(i)}} p_i(j).
\]

Thus, the total influence of node $j$ scales with the number of nodes $i$ such that $j \in \mathcal{N}^-(i)$ (i.e., its number of ancestors), \emph{modulated by the softmax weights} $p_i(j)$. In regimes where scores are not yet sharply separated (e.g., early training), $p_i(j)$ can be approximated as roughly uniform over the dominated set, yielding
\[
\frac{\partial \mathcal{L}_{\mathrm{graph}}}{\partial s_j}
\approx
\sum_{\substack{i \in V \\ j \in \mathcal{N}^-(i)}} 
\frac{1}{|\{i\} \cup \mathcal{N}^-(i)|},
\]
which makes the dependence on the number of ancestors explicit.

This induces a \emph{transitivity-aware weighting}: nodes that appear in many dominated sets (e.g., clearly suboptimal responses) receive larger aggregate gradient magnitude due to repeated inclusion across neighborhoods. At the same time, the softmax weights $p_i(j)$ adaptively modulate each contribution, preventing uniformly large updates. Together, this yields a curriculum-like effect in which clearly suboptimal responses are corrected more aggressively, while finer-grained distinctions among competitive samples are refined later through higher-confidence comparisons.

\paragraph{Layered graphs and groupwise weighting.}
Under the equivalence-class construction, nodes are partitioned into $G$ ordered groups. Let $g(i)$ denote the group index of node $i$. Then
\[
\mathcal{N}^-(i) = \bigcup_{g' > g(i)} \mathcal{G}_{g'}.
\]
In this case, all nodes within a group share identical dominated sets, and weighting reduces to a group-level structure.

\textbf{Corollary 1.}
\emph{
Under layered graphs, GraphDPO induces uniform weighting across all edges between any pair of groups, while assigning higher aggregate weight to comparisons involving lower-ranked groups.
}

\emph{Proof.}
Under the layered construction, every node $i \in \mathcal{C}_g$ has the same dominated set $\mathcal{N}^-(i) = \bigcup_{g' > g} \mathcal{C}_{g'}$ and hence the same choice set, so for a fixed dominated node $j$, the gradient contribution $p_i(j)$ is identical for all $i \in \mathcal{C}_g$: edges from any node of $\mathcal{C}_g$ to $j$ carry equal weight, and by symmetry over $j \in \mathcal{C}_h$, all edges between $\mathcal{C}_g$ and $\mathcal{C}_h$ are weighted uniformly. For the second claim, the ancestor set of $j \in \mathcal{C}_h$ is $\mathcal{A}(j) = \bigcup_{g < h} \mathcal{C}_g$, which is non-decreasing in $h$ under set inclusion; by Proposition~2, the aggregate weight on $j$ therefore increases with its group index, i.e., lower-ranked groups receive higher aggregate weight. \hfill $\square$

This explains why the method remains stable even when many samples share identical preference signals: within-group edges are absent, and the remaining constraints form a group-level structure with uniform per-edge weighting. It also yields an implicit curriculum: the worst responses, having the most ancestors, receive the strongest downward pressure, while near-optimal responses receive comparatively mild corrections.

\paragraph{Relation to multi-negative objectives.}
When the graph collapses to two groups (e.g., correct vs.\ incorrect), GraphDPO reduces to a group-wise softmax objective:
\[
\ell_i = -s_i + \log \left( \exp(s_i) + \sum_{j \in \text{neg}} \exp(s_j) \right),
\]
which resembles multi-negative DPO. However, the weighting structure differs: GraphDPO assigns shared normalization at the group level and avoids redundant intra-group comparisons, while listwise methods impose a strict ordering over all samples.

\paragraph{Robustness under noise and class imbalance.}
GraphDPO inherits robustness properties from log-sum-exp aggregation. Consider a node $i$ with dominated set $\mathcal{N}^-(i)$. The gradient contribution from any $j \in \mathcal{N}^-(i)$ is:
\[
\frac{\partial \ell_i}{\partial s_j}
=
\frac{\exp(s_j)}{\sum_{k \in \{i\} \cup \mathcal{N}^-(i)} \exp(s_k)},
\]
which downweights low-scoring or noisy samples.

\textbf{Proposition 3 (Adaptive weighting and noise attenuation).}
\emph{
The gradient contribution of a dominated sample is proportional to its softmax probability within the dominated set. As a result, samples with low scores (i.e., responses assigned low scores under the current model) receive exponentially smaller weight, while competitive samples receive larger updates.
}

\emph{Proof.}
From the gradient derivation above, the contribution of $j \in \mathcal{N}^-(i)$ to $\ell_i$ is exactly $p_i(j) = \exp(s_j)/\sum_{k \in \{i\} \cup \mathcal{N}^-(i)} \exp(s_k)$, the softmax probability of $j$ in the choice set of $i$, which establishes the first claim. For the second, fix two dominated samples $j, j' \in \mathcal{N}^-(i)$; their contributions satisfy $p_i(j)/p_i(j') = \exp(s_j - s_{j'})$, so a sample scored $\Delta$ below another receives weight smaller by the factor $e^{-\Delta}$, i.e., exponentially smaller in the score gap, while samples with scores close to $s_i$ retain weight bounded away from zero. \hfill $\square$

Under class imbalance, equivalence-class grouping prevents quadratic growth in constraints and avoids overcounting redundant comparisons within large classes.

\paragraph{Discussion.}
These results show that GraphDPO induces a structured, transitivity-aware reweighting of pairwise constraints that emphasizes globally consistent ordering while mitigating noise and redundancy. While this weighting is implicit, it aligns with the objective of learning coherent preference structures from multi-sample rollouts.

\section{Additional Results}\label{appd:additional_result}

\subsection{Preference-Layer Statistics and Depth-Stratified Analysis}
\label{appd:layer_stats}

We characterize the preference-graph structure per prompt at two levels: what the verifier \emph{can} support (achievable layers, determined by the verifier and stable during training) and what the policy's rollouts \emph{actually realize} (realized layers, measured from base-model rollouts at $K{=}8$).

\paragraph{Achievable layers.}
For binary verifiers, the number of achievable layers is 2; for APPS it is $\min(\text{number of hidden test groups}, 10)+1$ distinct pass-rate levels:

\begin{table}[ht]
\centering
\caption{Achievable preference layers per prompt, determined by verifier structure.}
\label{tab:achievable_layers}
\small
\setlength{\tabcolsep}{4pt}
\begin{tabular}{lccc}
\toprule
Benchmark & Verifier & Achievable layers/prompt & \% prompts $\ge$3 layers \\
\midrule
GSM8K & binary exact-match & 2 & 0 \\
MATH  & binary exact-match & 2 & 0 \\
APPS  & fraction of tests passed & median 11 (mean 7.9, max 11) & 100 \\
\bottomrule
\end{tabular}
\end{table}

APPS supports deep graphs on all prompts: every prompt admits $\ge$3 ordered levels and 84\% admit $\ge$4. On GSM8K and MATH the realized graph is always two-layer, since binary correctness admits no intermediate levels regardless of $K$; there, GraphDPO reduces to a grouped correct-vs-incorrect objective with tie masking (Appendix~\ref{appd:opt_properties}), and the gains over MPO/SWEPO come from per-response dominated-set normalization rather than graph depth.

\paragraph{Realized layers.}
The graph the model trains on depends on which levels the policy's rollouts hit. From base-model rollouts at $K{=}8$ on the APPS test set (400 prompts), 17.8\% of prompts realize $\ge$3 layers and 47\% realize $\ge$2.

\paragraph{Depth-stratified accuracy.}
Stratifying APPS prompts by realized layer count, GraphDPO's improvement over the base policy on the same prompts grows monotonically with graph depth:

\begin{table}[ht]
\centering
\caption{APPS improvement over the base policy, stratified by realized preference-graph depth.}
\label{tab:depth_stratified}
\begin{tabular}{lcc}
\toprule
Realized layers & GraphDPO $-$ base (APPS pass-rate) & $n$ \\
\midrule
2 layers       & $+16.8$ & 329 \\
3 layers       & $+25.9$ & 58 \\
$\ge 4$ layers & $+36.2$ & 13 \\
\bottomrule
\end{tabular}
\end{table}

The margin more than doubles from 2-layer to $\ge$4-layer prompts ($+16.8 \to +36.2$), indicating that the improvement is attributable to the multi-layer graph structure the reward induces, not merely to a denser reward. Consistently, APPS is the benchmark where GraphDPO's overall margin over the strongest baseline is largest (Table~\ref{tab:main_results}).

\subsection{Baseline \texorpdfstring{$K$}{K}-Scaling and Anchoring Isolation}
\label{appd:kscaling}

\paragraph{Baseline $K$-scaling.}
To separate the value of the graph objective from the benefit of using more rollouts, we train PRO and GraphDPO through the same pipeline at $K = 2, 4, 8$ on GSM8K (single matched seed for a controlled comparison):

\begin{table}[ht]
\centering
\caption{GSM8K accuracy vs.\ rollout count $K$ (PRO vs.\ GraphDPO, same pipeline, single matched seed).}
\label{tab:kscaling}
\begin{tabular}{lcc}
\toprule
$K$ & PRO & GraphDPO \\
\midrule
2 & 86.7 & 88.78 \\
4 & 88.3 & 92.04 \\
8 & 88.7 & 92.42 \\
\bottomrule
\end{tabular}
\end{table}

The gap grows from $+2.08$ at $K{=}2$ to $+3.74$ at $K{=}4$ and is maintained at $K{=}8$ ($+3.72$): PRO with the same rollout budget cannot close the gap at any $K$, so GraphDPO's advantage is not an artifact of rollout count. The multi-seed means in Table~\ref{tab:main_results} confirm the same pattern at $K{=}8$ ($92.42 - 88.82 = +3.60$).

\paragraph{Anchoring isolation.}
To test whether the ground-truth anchor alone explains GraphDPO's advantage, we add the same gold trajectory to PRO (GSM8K, $K{=}8$, same seed): PRO achieves 88.7 and PRO$+$GT achieves 88.3. The oracle trajectory gives PRO no gain (within noise), confirming that GraphDPO's advantage comes from the graph-structured objective rather than from the anchor. Together with the ablation in Table~\ref{tab:gsm8k_ablation}, where the anchoring gain shrinks from $+3.03$ at $K{=}2$ to $+0.33$ at $K{=}8$, this supports presenting anchoring as a low-$K$ stabilizer whose benefit a richer graph largely subsumes.

\subsection{Worked Example}
\label{appd:worked_example}

We illustrate the construction on a real APPS prompt with $K{=}8$ rollouts whose test-pass fractions are $[1.0,\, 1.0,\, 0.6,\, 0.6,\, 0.6,\, 0.0,\, 0.0,\, 0.0]$. The pass fractions induce three equivalence classes: $\mathcal{C}_1 = \{y_1, y_2\}$ (pass rate 1.0), $\mathcal{C}_2 = \{y_3, y_4, y_5\}$ (0.6), and $\mathcal{C}_3 = \{y_6, y_7, y_8\}$ (0.0), ordered $\mathcal{C}_1 \succ \mathcal{C}_2 \succ \mathcal{C}_3$. Edges connect every node to every node in strictly lower classes ($2{\times}3 + 2{\times}3 + 3{\times}3 = 21$ edges), while the pairs within each class, including responses passing different test subsets at the same rate, receive no edge and exchange no gradient. Each node's loss normalizes over itself and its dominated classes: for example, $y_3$'s choice set is $\{y_3\} \cup \mathcal{C}_3$, and $y_1$'s is $\{y_1\} \cup \mathcal{C}_2 \cup \mathcal{C}_3$. The three-pass algorithm of Appendix~\ref{appd:complexity} computes all eight losses with a single suffix scan over the three class sums. By contrast, a binary GSM8K prompt with 3 correct and 5 incorrect rollouts yields only two classes and a single class boundary; each correct response normalizes over itself and the 5 incorrect ones, and no gradient flows among the 3 correct responses.

\subsection{Sensitivity to Ground-Truth Anchoring Weight}
\label{appd:lambda_sensitivity}

\begin{figure}[!t]\centering
\includegraphics[width=0.75\linewidth]{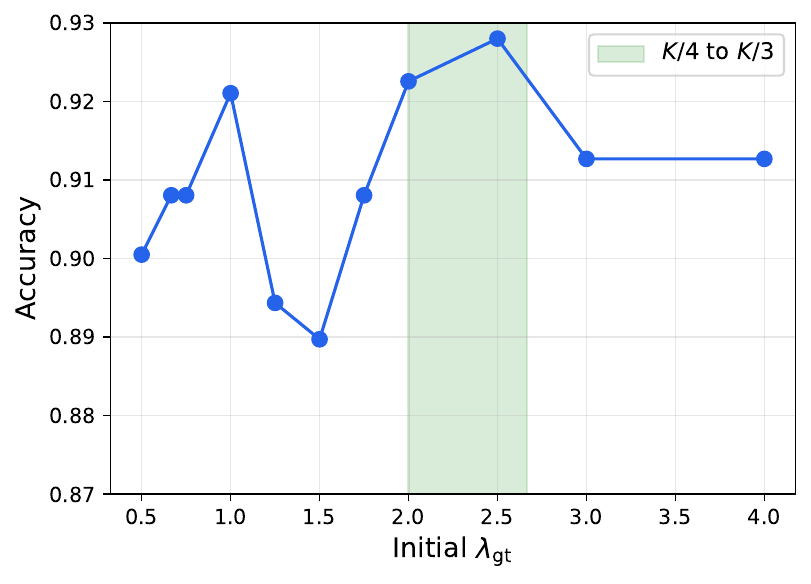}
 \caption{Sensitivity of GraphDPO to the initial anchoring weight $\lambda_{\mathrm{gt}}$ on GSM8K. The shaded region indicates the recommended range.}
 \label{fig:lambda_sensitivity_gsm8k}
\end{figure}

We study the sensitivity of GraphDPO to the initial ground-truth anchoring weight $\lambda_{\mathrm{gt}}$. Figure~\ref{fig:lambda_sensitivity_gsm8k} shows GSM8K accuracy as a function of the initial value, with a linear decay schedule applied during training. We observe that performance is stable across a broad range of values, with the best results achieved for $\lambda_{\mathrm{gt}} \in [2.0, 2.5]$. Extremely small values underutilize the anchoring signal, while overly large values slightly degrade performance, likely due to over-constraining the policy early in training. These results indicate that GraphDPO is not overly sensitive to the exact choice of anchoring strength, and that a moderate initial value provides a good balance between stability and flexibility.

\paragraph{Rationale for the $[K/4, K/3]$ initialization.}
The recommended range arises from gradient-magnitude matching between the anchor term and the graph term. Both terms share the same log-sum-exp scale, but the graph term aggregates over all nodes with non-empty dominated sets, whereas the anchor term is a single node's loss. For the single gold node to contribute on the order of one group's worth of gradient early in training, its weight should be comparable to the expected number of dominated groups, which scales with $K$. At $K{=}8$, $K/4 = 2.0$ and $K/3 \approx 2.67$, matching the empirically optimal range $[2.0, 2.5]$ (with less than $0.2\%$ variation across it). We therefore view the initialization as a robust heuristic derived from gradient-magnitude matching rather than a hand-tuned parameter. The time-dependent decay follows the same logic in reverse: as the policy improves and more rollouts achieve correctness, the graph supplies increasing positive signal, and $\lambda_{\mathrm{gt}}$ anneals toward 1 (equal weight with other responses) so that the anchor does not dominate the objective at convergence. This is consistent with the empirical finding that anchoring's marginal value decreases with $K$ (Table~\ref{tab:gsm8k_ablation}): a richer graph subsumes the anchor's role.

%%%%%%%%%%%%%%%%%%%%%%%%%%%%%%%%%%%%%%%%%%%%%%%%%%%%%%%%%%%%

\newpage
\section*{NeurIPS Paper Checklist}

\begin{enumerate}

\item {\bf Claims}
    \item[] Question: Do the main claims made in the abstract and introduction accurately reflect the paper's contributions and scope?
    \item[] Answer: \answerYes{}
    \item[] Justification: The abstract and introduction clearly state the key contributions (i.e., graph-structured preference modeling, equivalence-class handling, linear-time learning under the layered construction, and stabilizing ground-truth anchoring), state the scope of each claim (the $O(K)$ bound applies to the layered construction; experiments use ranked equivalence classes with ties and incomparabilities), and these are directly supported by the theoretical formulation in Section 4 and empirical results in Section 5.

\item {\bf Limitations}
    \item[] Question: Does the paper discuss the limitations of the work performed by the authors?
    \item[] Answer: \answerYes{}
    \item[] Justification: The discussion on the limitations of the work is provided at the end of the Conclusion section.

\item {\bf Theory assumptions and proofs}
    \item[] Question: For each theoretical result, does the paper provide the full set of assumptions and a complete (and correct) proof?
    \item[] Answer: \answerYes{}
    \item[] Justification: Appendix~\ref{appd:opt_properties} presents Propositions 1--3 and Corollary~1 with clearly stated assumptions and complete proofs characterizing pairwise consistency, implicit edge weighting, and noise attenuation properties of the GraphDPO objective; Appendix~\ref{appd:complexity} provides the explicit algorithm and complexity lemma supporting the $O(K)$ claim.

    \item {\bf Experimental result reproducibility}
    \item[] Question: Does the paper fully disclose all the information needed to reproduce the main experimental results of the paper to the extent that it affects the main claims and/or conclusions of the paper (regardless of whether the code and data are provided or not)?
    \item[] Answer: \answerYes{}
    \item[] Justification: Appendix~\ref{appd:exp_setting} provides detailed descriptions of datasets, preprocessing, rollout generation, model configurations, hyperparameters, and training setup, including optimizer, learning rates, batch sizes, and hardware.

\item {\bf Open access to data and code}
    \item[] Question: Does the paper provide open access to the data and code, with sufficient instructions to faithfully reproduce the main experimental results, as described in supplemental material?
    \item[] Answer: \answerYes{}
    \item[] Justification: GraphDPO is implemented on top of the open-source verl framework~\citep{sheng2024hybridflow} and changes only the policy update; Appendix~\ref{appd:code} lists all modifications and gives the complete loss implementation, and Appendices~\ref{appd:exp_setting} and~\ref{appd:complexity} give data preparation, hyperparameters, and algorithmic details needed to reproduce the main results. All datasets used are publicly available.

\item {\bf Experimental setting/details}
    \item[] Question: Does the paper specify all the training and test details (e.g., data splits, hyperparameters, how they were chosen, type of optimizer) necessary to understand the results?
    \item[] Answer: \answerYes{}
    \item[] Justification: The paper and Appendix A specify datasets, preprocessing, model configurations, hyperparameters, and training procedures in sufficient detail for reproducibility.

\item {\bf Experiment statistical significance}
    \item[] Question: Does the paper report error bars suitably and correctly defined or other appropriate information about the statistical significance of the experiments?
    \item[] Answer: \answerYes{}
    \item[] Justification: We report mean $\pm$ standard deviation over multiple independent training runs (with different random seeds) for GraphDPO and all preference-based baselines in Table~\ref{tab:main_results}, which constitute the primary results supporting our claims. The reported error bars capture variability due to stochastic training dynamics, including initialization and sampling. All methods are evaluated under identical settings (datasets, rollout budgets, and training configurations), and the performance gaps between GraphDPO and the strongest baselines substantially exceed the observed seed-level variance (e.g., the GSM8K gap over SWEPO is roughly 14$\times$ the pooled standard deviation). We explicitly indicate that the reported uncertainty corresponds to one standard deviation across runs.

\item {\bf Experiments compute resources}
    \item[] Question: For each experiment, does the paper provide sufficient information on the computer resources (type of compute workers, memory, time of execution) needed to reproduce the experiments?
    \item[] Answer: \answerYes{}
    \item[] Justification: Appendix A describes the hardware (e.g., GPU type and memory), training setup, and approximate compute requirements for the experiments.
    
\item {\bf Code of ethics}
    \item[] Question: Does the research conducted in the paper conform, in every respect, with the NeurIPS Code of Ethics \url{https://neurips.cc/public/EthicsGuidelines}?
    \item[] Answer: \answerYes{}
    \item[] Justification: The research complies with the NeurIPS Code of Ethics and does not involve any known ethical violations.

\item {\bf Broader impacts}
    \item[] Question: Does the paper discuss both potential positive societal impacts and negative societal impacts of the work performed?
    \item[] Answer: \answerYes{}
    \item[] Justification: The paper discusses both positive impacts (improved reliability of LLMs) and potential negative impacts (bias from preference signals) in the Conclusion section.
    
\item {\bf Safeguards}
    \item[] Question: Does the paper describe safeguards that have been put in place for responsible release of data or models that have a high risk for misuse (e.g., pre-trained language models, image generators, or scraped datasets)?
    \item[] Answer: \answerNA{}
    \item[] Justification: The paper does not involve models or datasets that pose significant risk of misuse.

\item {\bf Licenses for existing assets}
    \item[] Question: Are the creators or original owners of assets (e.g., code, data, models), used in the paper, properly credited and are the license and terms of use explicitly mentioned and properly respected?
    \item[] Answer: \answerYes{}
    \item[] Justification: All datasets used (e.g., GSM8K, MATH, APPS) are publicly available and properly cited.

\item {\bf New assets}
    \item[] Question: Are new assets introduced in the paper well documented and is the documentation provided alongside the assets?
    \item[] Answer: \answerNA{}
    \item[] Justification: The paper does not introduce or release new datasets, models, or other assets.

\item {\bf Crowdsourcing and research with human subjects}
    \item[] Question: For crowdsourcing experiments and research with human subjects, does the paper include the full text of instructions given to participants and screenshots, if applicable, as well as details about compensation (if any)? 
    \item[] Answer: \answerNA{}
    \item[] Justification: The paper does not involve crowdsourcing nor research with human subjects.

\item {\bf Institutional review board (IRB) approvals or equivalent for research with human subjects}
    \item[] Question: Does the paper describe potential risks incurred by study participants, whether such risks were disclosed to the subjects, and whether Institutional Review Board (IRB) approvals (or an equivalent approval/review based on the requirements of your country or institution) were obtained?
    \item[] Answer: \answerNA{}
    \item[] Justification: The paper does not involve crowdsourcing nor research with human subjects.

\item {\bf Declaration of LLM usage}
    \item[] Question: Does the paper describe the usage of LLMs if it is an important, original, or non-standard component of the core methods in this research? Note that if the LLM is used only for writing, editing, or formatting purposes and does \emph{not} impact the core methodology, scientific rigor, or originality of the research, declaration is not required.
    %this research? 
    \item[] Answer: \answerYes{}
    \item[] Justification: Large language models are used as base models and for rollout generation, which are core components of the proposed method and experimental setup.

\end{enumerate}


\begin{thebibliography}{52}
\providecommand{\natexlab}[1]{#1}
\providecommand{\url}[1]{\texttt{#1}}
\expandafter\ifx\csname urlstyle\endcsname\relax
  \providecommand{\doi}[1]{doi: #1}\else
  \providecommand{\doi}{doi: \begingroup \urlstyle{rm}\Url}\fi

\bibitem[Bai et~al.(2022)]{bai2022constitutional}
Yuntao Bai et~al.
\newblock Constitutional ai: Harmlessness from ai feedback.
\newblock \emph{arXiv preprint arXiv:2212.08073}, 2022.

\bibitem[Ganguli et~al.(2022)]{ganguli2022predictability}
Deep Ganguli et~al.
\newblock Predictability and surprise in large generative models.
\newblock \emph{arXiv preprint arXiv:2202.07785}, 2022.

\bibitem[Touvron et~al.(2023)]{touvron2023llama}
Hugo Touvron et~al.
\newblock Llama: Open and efficient foundation language models.
\newblock \emph{arXiv preprint arXiv:2302.13971}, 2023.

\bibitem[Christiano et~al.(2017)Christiano, Leike, Brown, Martic, Legg, and Amodei]{christiano2017deep}
Paul~F Christiano, Jan Leike, Tom~B Brown, Miljan Martic, Shane Legg, and Dario Amodei.
\newblock Deep reinforcement learning from human preferences.
\newblock In \emph{NeurIPS}, 2017.

\bibitem[Stiennon et~al.(2020)Stiennon, Ouyang, Wu, Ziegler, Lowe, Voss, Radford, Amodei, and Christiano]{stiennon2020learning}
Nisan Stiennon, Long Ouyang, Jeff Wu, Daniel Ziegler, Ryan Lowe, Chelsea Voss, Alec Radford, Dario Amodei, and Paul Christiano.
\newblock Learning to summarize with human feedback.
\newblock In \emph{NeurIPS}, 2020.

\bibitem[Ouyang et~al.(2022)Ouyang, Wu, Jiang, Almeida, Wainwright, et~al.]{ouyang2022training}
Long Ouyang, Jeffrey Wu, Xu~Jiang, Diogo Almeida, Carroll Wainwright, et~al.
\newblock Training language models to follow instructions with human feedback.
\newblock \emph{Advances in Neural Information Processing Systems}, 35, 2022.

\bibitem[Schulman et~al.(2017)Schulman, Wolski, Dhariwal, Radford, and Klimov]{schulman2017ppo}
John Schulman, Filip Wolski, Prafulla Dhariwal, Alec Radford, and Oleg Klimov.
\newblock Proximal policy optimization algorithms.
\newblock \emph{arXiv preprint arXiv:1707.06347}, 2017.

\bibitem[Ziegler et~al.(2019)Ziegler, Stiennon, Wu, Brown, Radford, Amodei, and Christiano]{ziegler2019fine}
Daniel~M. Ziegler, Nisan Stiennon, Jeffrey Wu, Tom Brown, Alec Radford, Dario Amodei, and Paul Christiano.
\newblock Fine-tuning language models from human preferences.
\newblock \emph{arXiv preprint arXiv:1909.08593}, 2019.

\bibitem[Rafailov et~al.(2023)Rafailov, Sharma, Mitchell, Ermon, Manning, and Finn]{rafailov2023direct}
Rafael Rafailov, Archit Sharma, Eric Mitchell, Stefano Ermon, Christopher~D. Manning, and Chelsea Finn.
\newblock Direct preference optimization: Your language model is secretly a reward model.
\newblock \emph{Advances in Neural Information Processing Systems}, 36, 2023.

\bibitem[Li et~al.(2023)Li, Zhang, Dubois, Taori, Gulrajani, Guestrin, Liang, and Hashimoto]{li2023alpacaeval}
Xuechen Li, Tianyi Zhang, Yann Dubois, Rohan Taori, Ishaan Gulrajani, Carlos Guestrin, Percy Liang, and Tatsunori~B Hashimoto.
\newblock Alpacaeval: An automatic evaluator of instruction-following models, 2023.

\bibitem[Azar et~al.(2024)Azar, Guo, Piot, Munos, Rowland, Valko, and Calandriello]{azar2024general}
Mohammad~Gheshlaghi Azar, Zhaohan~Daniel Guo, Bilal Piot, Remi Munos, Mark Rowland, Michal Valko, and Daniele Calandriello.
\newblock A general theoretical paradigm to understand learning from human preferences.
\newblock In \emph{International Conference on Artificial Intelligence and Statistics}, pages 4447--4455. PMLR, 2024.

\bibitem[Ethayarajh et~al.(2024)]{ethayarajh2024kto}
Kawin Ethayarajh et~al.
\newblock Kto: Model alignment as prospect theoretic optimization.
\newblock \emph{arXiv preprint arXiv:2402.01306}, 2024.

\bibitem[Meng et~al.(2024)Meng, Xia, and Chen]{meng2024simpo}
Yu~Meng, Mengzhou Xia, and Danqi Chen.
\newblock Simpo: Simple preference optimization with a reference-free reward.
\newblock \emph{arXiv preprint arXiv:2405.14734}, 2024.

\bibitem[Hong et~al.(2024)Hong, Lee, and Thorne]{hong2024orpo}
Jiwoo Hong, Noah Lee, and James Thorne.
\newblock Orpo: Monolithic preference optimization without reference model.
\newblock In \emph{Proceedings of the 2024 Conference on Empirical Methods in Natural Language Processing}, pages 11170--11189, 2024.

\bibitem[Chen et~al.(2024{\natexlab{a}})Chen, Deng, Yuan, Ji, and Gu]{chen2024spin}
Zixiang Chen, Yihe Deng, Huizhuo Yuan, Kaixuan Ji, and Quanquan Gu.
\newblock Self-play fine-tuning converts weak language models to strong language models.
\newblock \emph{arXiv preprint arXiv:2401.01335}, 2024{\natexlab{a}}.

\bibitem[Hendrycks et~al.(2021{\natexlab{a}})Hendrycks, Basart, Kadavath, Mazeika, Arora, Guo, Burns, Puranik, He, Song, et~al.]{hendrycks2021apps}
Dan Hendrycks, Steven Basart, Saurav Kadavath, Mantas Mazeika, Akul Arora, Ethan Guo, Collin Burns, Samir Puranik, Horace He, Dawn Song, et~al.
\newblock Measuring coding challenge competence with apps.
\newblock \emph{arXiv preprint arXiv:2105.09938}, 2021{\natexlab{a}}.

\bibitem[Cobbe et~al.(2021)Cobbe, Kosaraju, Bavarian, Chen, Jun, Kaiser, Plappert, Tworek, Hilton, Nakano, et~al.]{cobbe2021training}
Karl Cobbe, Vineet Kosaraju, Mohammad Bavarian, Mark Chen, Heewoo Jun, Lukasz Kaiser, Matthias Plappert, Jerry Tworek, Jacob Hilton, Reiichiro Nakano, et~al.
\newblock Training verifiers to solve math word problems.
\newblock \emph{arXiv preprint arXiv:2110.14168}, 2021.

\bibitem[Hendrycks et~al.(2021{\natexlab{b}})Hendrycks, Burns, Kadavath, Arora, Basart, Tang, Song, and Steinhardt]{hendrycks2021measuring}
Dan Hendrycks, Collin Burns, Saurav Kadavath, Akul Arora, Steven Basart, Eric Tang, Dawn Song, and Jacob Steinhardt.
\newblock Measuring mathematical problem solving with the math dataset.
\newblock \emph{arXiv preprint arXiv:2103.03874}, 2021{\natexlab{b}}.

\bibitem[Shao et~al.(2024)Shao, Wang, Zhu, Xu, Song, Zhang, Li, Wu, and Guo]{shao2024deepseekmath}
Zhihong Shao, Peiyi Wang, Qihao Zhu, Runxin Xu, Junxiao Song, Mingchuan Zhang, YK~Li, Y~Wu, and Daya Guo.
\newblock Deepseekmath: Pushing the limits of mathematical reasoning in open language models.
\newblock \emph{arXiv preprint arXiv:2402.03300}, 2024.

\bibitem[Lightman et~al.(2023)Lightman, Kosaraju, Burda, Edwards, Baker, Lee, Leike, Schulman, Sutskever, and Cobbe]{lightman2023lets}
Hunter Lightman, Vineet Kosaraju, Yuri Burda, Harrison Edwards, Bowen Baker, Teddy Lee, Jan Leike, John Schulman, Ilya Sutskever, and Karl Cobbe.
\newblock Let's verify step by step.
\newblock In \emph{The twelfth international conference on learning representations}, 2023.

\bibitem[Wang et~al.(2022)Wang, Wei, Schuurmans, Le, Chi, Narang, Chowdhery, and Zhou]{wang2022self}
Xuezhi Wang, Jason Wei, Dale Schuurmans, Quoc Le, Ed~Chi, Sharan Narang, Aakanksha Chowdhery, and Denny Zhou.
\newblock Self-consistency improves chain of thought reasoning in language models.
\newblock \emph{arXiv preprint arXiv:2203.11171}, 2022.

\bibitem[Liu(2026)]{liu2026jury}
Ning Liu.
\newblock {LLMs} as a jury: Cross-model consensus can outperform process reward models for {LLM} reasoning.
\newblock \emph{arXiv preprint arXiv:2607.10139}, 2026.

\bibitem[Vinyals et~al.(2015)Vinyals, Bengio, and Kudlur]{vinyals2016order}
Oriol Vinyals, Samy Bengio, and Manjunath Kudlur.
\newblock Order matters: Sequence to sequence for sets.
\newblock \emph{arXiv preprint arXiv:1511.06391}, 2015.

\bibitem[Cao et~al.(2007)Cao, Qin, Liu, Tsai, and Li]{cao2007learning}
Zhe Cao, Tao Qin, Tie-Yan Liu, Ming-Feng Tsai, and Hang Li.
\newblock Learning to rank: from pairwise approach to listwise approach.
\newblock In \emph{Proceedings of the 24th international conference on Machine learning}, pages 129--136, 2007.

\bibitem[Song et~al.(2024)Song, Yu, Li, Yu, Huang, Li, and Wang]{song2024preference}
Feifan Song, Bowen Yu, Minghao Li, Haiyang Yu, Fei Huang, Yongbin Li, and Houfeng Wang.
\newblock Preference ranking optimization for human alignment.
\newblock In \emph{Proceedings of the AAAI Conference on Artificial Intelligence}, volume~38, pages 18990--18998, 2024.

\bibitem[Liu et~al.(2025)Liu, Qin, Wu, Shen, Khalman, Joshi, Zhao, Saleh, Baumgartner, Liu, et~al.]{liu2025lipo}
Tianqi Liu, Zhen Qin, Junru Wu, Jiaming Shen, Misha Khalman, Rishabh Joshi, Yao Zhao, Mohammad Saleh, Simon Baumgartner, Jialu Liu, et~al.
\newblock Lipo: Listwise preference optimization through learning-to-rank.
\newblock In \emph{Proceedings of the 2025 Conference of the Nations of the Americas Chapter of the Association for Computational Linguistics: Human Language Technologies (Volume 1: Long Papers)}, pages 2404--2420, 2025.

\bibitem[Plackett(1975)]{plackett1975analysis}
Robin~L Plackett.
\newblock The analysis of permutations.
\newblock \emph{Journal of the Royal Statistical Society Series C: Applied Statistics}, 24\penalty0 (2):\penalty0 193--202, 1975.

\bibitem[Luce et~al.(1959)]{luce1959individual}
R~Duncan Luce et~al.
\newblock \emph{Individual choice behavior}, volume~4.
\newblock Wiley New York, 1959.

\bibitem[Burges(2010)]{burges2010ranknet}
Christopher~JC Burges.
\newblock From ranknet to lambdarank to lambdamart: An overview.
\newblock \emph{Learning}, 11\penalty0 (23-581):\penalty0 81, 2010.

\bibitem[Liu(2009)]{liu2009learning}
Tie-Yan Liu.
\newblock Learning to rank for information retrieval.
\newblock \emph{Foundations and Trends{\textregistered} in Information Retrieval}, 3\penalty0 (3):\penalty0 225--331, 2009.

\bibitem[Wang et~al.(2024)Wang, Zheng, Li, Zhang, Gui, and Liu]{wang2024rescue}
Yikun Wang, Rui Zheng, Haoming Li, Qi~Zhang, Tao Gui, and Fei Liu.
\newblock Rescue: Ranking llm responses with partial ordering to improve response generation.
\newblock In \emph{Proceedings of the 62nd Annual Meeting of the Association for Computational Linguistics (Volume 4: Student Research Workshop)}, pages 261--272, 2024.

\bibitem[Chen et~al.(2024{\natexlab{b}})Chen, Malladi, Zhang, Chen, Zhang, Ranganath, and Cho]{chen2024preference}
Angelica Chen, Sadhika Malladi, Lily~H Zhang, Xinyi Chen, Qiuyi Zhang, Rajesh Ranganath, and Kyunghyun Cho.
\newblock Preference learning algorithms do not learn preference rankings.
\newblock \emph{Advances in Neural Information Processing Systems}, 37:\penalty0 101928--101968, 2024{\natexlab{b}}.

\bibitem[Koller and Friedman(2009)]{koller2009probabilistic}
Daphne Koller and Nir Friedman.
\newblock \emph{Probabilistic graphical models: principles and techniques}.
\newblock MIT press, 2009.

\bibitem[Yuan et~al.(2023)Yuan, Yuan, Tan, Wang, Huang, and Huang]{yuan2023rrhf}
Hongyi Yuan, Zheng Yuan, Chuanqi Tan, Wei Wang, Songfang Huang, and Fei Huang.
\newblock Rrhf: Rank responses to align language models with human feedback.
\newblock \emph{Advances in Neural Information Processing Systems}, 36:\penalty0 10935--10950, 2023.

\bibitem[Najafi and Fyshe(2026)]{najafi2026offline}
Saeed Najafi and Alona Fyshe.
\newblock Offline preference optimization via maximum marginal likelihood estimation.
\newblock In \emph{Proceedings of the 19th Conference of the European Chapter of the Association for Computational Linguistics (Volume 1: Long Papers)}, pages 6751--6764, 2026.

\bibitem[Kim et~al.(2024)Kim, Seo, Liu, Shin, and Lee]{kim2024margin}
Kyuyoung Kim, Ah~Jeong Seo, Hao Liu, Jinwoo Shin, and Kimin Lee.
\newblock Margin matching preference optimization: Enhanced model alignment with granular feedback.
\newblock In \emph{Findings of the Association for Computational Linguistics: EMNLP 2024}, pages 13554--13570, 2024.

\bibitem[Amini et~al.(2024)Amini, Vieira, and Cotterell]{amini2024direct}
Afra Amini, Tim Vieira, and Ryan Cotterell.
\newblock Direct preference optimization with an offset.
\newblock In \emph{Findings of the Association for Computational Linguistics: ACL 2024}, pages 9954--9972, 2024.

\bibitem[Pang et~al.(2026)Pang, Zhu, Di, Zhang, Wang, Qian, and Liu]{pang2026small}
Jinlong Pang, Zhaowei Zhu, Na~Di, Yichi Zhang, Yaxuan Wang, Chen Qian, and Yang Liu.
\newblock Small-margin preferences still matter-if you train them right.
\newblock \emph{arXiv preprint arXiv:2602.00954}, 2026.

\bibitem[Sun et~al.(2024)Sun, Haider, Zhang, Yang, Qiu, Yin, Wang, Bartlett, and Zanette]{sun2024fast}
Hanshi Sun, Momin Haider, Ruiqi Zhang, Huitao Yang, Jiahao Qiu, Ming Yin, Mengdi Wang, Peter Bartlett, and Andrea Zanette.
\newblock Fast best-of-n decoding via speculative rejection.
\newblock \emph{Advances in Neural Information Processing Systems}, 37:\penalty0 32630--32652, 2024.

\bibitem[Nakano et~al.(2021)]{nakano2021webgpt}
Reiichiro Nakano et~al.
\newblock Webgpt: Browser-assisted question-answering with human feedback.
\newblock \emph{arXiv preprint arXiv:2112.09332}, 2021.

\bibitem[Gupta et~al.(2024)Gupta, Madhavan, Zhang, Natarajan, Bansal, and Rajmohan]{gupta2025mpo}
Taneesh Gupta, Rahul Madhavan, Xuchao Zhang, Nagarajan Natarajan, Chetan Bansal, and Saravan Rajmohan.
\newblock Multi-preference optimization: Generalizing dpo via set-level contrasts.
\newblock \emph{arXiv preprint arXiv:2412.04628}, 2024.

\bibitem[Gupta et~al.(2025)Gupta, Madhavan, Zhang, Bansal, and Rajmohan]{gupta2025swepo}
Taneesh Gupta, Rahul Madhavan, Xuchao Zhang, Chetan Bansal, and Saravan Rajmohan.
\newblock Swepo: Simultaneous weighted preference optimization for group contrastive alignment.
\newblock In \emph{ICLR 2025 Workshop on Bidirectional Human-AI Alignment}, 2025.

\bibitem[Yin et~al.(2024)Yin, Wang, Gu, Huang, Chen, and Zhou]{yin2024relative}
Yueqin Yin, Zhendong Wang, Yi~Gu, Hai Huang, Weizhu Chen, and Mingyuan Zhou.
\newblock Relative preference optimization: Enhancing llm alignment through contrasting responses across identical and diverse prompts.
\newblock \emph{arXiv preprint arXiv:2402.10958}, 2024.

\bibitem[Wang et~al.(2018)Wang, Li, Golbandi, Bendersky, and Najork]{wang2018lambdaloss}
Xuanhui Wang, Cheng Li, Nadav Golbandi, Michael Bendersky, and Marc Najork.
\newblock The lambdaloss framework for ranking metric optimization.
\newblock In \emph{Proceedings of the 27th ACM international conference on information and knowledge management}, pages 1313--1322, 2018.

\bibitem[Guiver and Snelson(2009)]{guiver2009bayesian}
John Guiver and Edward Snelson.
\newblock Bayesian inference for plackett--luce ranking models.
\newblock \emph{Proceedings of the 26th International Conference on Machine Learning}, 2009.

\bibitem[Dehghani et~al.(2017)Dehghani, Zamani, Severyn, Kamps, and Croft]{dehghani2017neural}
Mostafa Dehghani, Hamed Zamani, Aliaksei Severyn, Jaap Kamps, and W~Bruce Croft.
\newblock Neural ranking models with weak supervision.
\newblock In \emph{Proceedings of the 40th international ACM SIGIR conference on research and development in information retrieval}, pages 65--74, 2017.

\bibitem[Kipf and Welling(2017)]{kipf2017gcn}
Thomas Kipf and Max Welling.
\newblock Semi-supervised classification with graph convolutional networks.
\newblock In \emph{ICLR}, 2017.

\bibitem[Veli{\v{c}}kovi{\'c} et~al.(2018)]{velivckovic2018gat}
Petar Veli{\v{c}}kovi{\'c} et~al.
\newblock Graph attention networks.
\newblock In \emph{ICLR}, 2018.

\bibitem[Sekhari et~al.(2023)Sekhari, Sridharan, Sun, and Wu]{sekhari2023contextual}
Ayush Sekhari, Karthik Sridharan, Wen Sun, and Runzhe Wu.
\newblock Contextual bandits and imitation learning with preference-based active queries.
\newblock \emph{Advances in Neural Information Processing Systems}, 36:\penalty0 11261--11295, 2023.

\bibitem[Hu et~al.(2026)Hu, Zhang, Xiong, Ratner, Ding, and Krishna]{hu2026towards}
Zhengyu Hu, Jieyu Zhang, Zhihan Xiong, Alexander Ratner, Kaize Ding, and Ranjay Krishna.
\newblock Towards acyclic preference evaluation of language models via multiple evaluators.
\newblock In \emph{Proceedings of the AAAI Conference on Artificial Intelligence}, volume~40, pages 21903--21911, 2026.

\bibitem[Liu et~al.(2026)Liu, Sreekar, Kujanp{\"a}{\"a}, Zhu, Liu, Sun, Menendez, Bales, Yang, Alam, Yu, Liu, Klinkner, and Malmasi]{liu2026eluna}
Ning Liu, P~Aditya Sreekar, Kalle Kujanp{\"a}{\"a}, Zhaoxuan Zhu, Kaiwen Liu, Chuanneng Sun, Jorge~Marchena Menendez, Matthew Bales, Tianyu Yang, Shahnawaz Alam, Rose Yu, Baoyuan Liu, Kristina Klinkner, and Shervin Malmasi.
\newblock Eluna: An agentic {LLM} system for automating warehouse operations with reasoning and task execution.
\newblock \emph{arXiv preprint arXiv:2607.08960}, 2026.

\bibitem[Sheng et~al.(2024)Sheng, Zhang, Ye, Wu, Zhang, Zhang, Peng, Lin, and Wu]{sheng2024hybridflow}
Guangming Sheng, Chi Zhang, Zilingfeng Ye, Xibin Wu, Wang Zhang, Ru~Zhang, Yanghua Peng, Haibin Lin, and Chuan Wu.
\newblock Hybridflow: A flexible and efficient rlhf framework.
\newblock \emph{arXiv preprint arXiv:2409.19256}, 2024.

\end{thebibliography}
\end{document}